\documentclass[lettersize,journal]{IEEEtran}
\usepackage[utf8]{inputenc} 
\usepackage[T1]{fontenc}    

\usepackage{amsmath,amsfonts,amssymb} 

\usepackage{algorithmicx} 
\usepackage{algpseudocode} 
\usepackage{algorithm}     
\usepackage{needspace}
\usepackage{array}     
\usepackage{graphicx}  
\usepackage{booktabs}  
\usepackage[caption=false,font=normalsize,labelfont=sf,textfont=sf]{subfig} 
\usepackage{stfloats}  

\usepackage{textcomp} 
\usepackage{url}      
\usepackage{verbatim} 
\usepackage{cite}     
\usepackage{balance}  

\usepackage{xcolor} 
\definecolor{cvprblue}{rgb}{0.21,0.49,0.74} 

\usepackage[pagebackref,breaklinks,colorlinks,allcolors=cvprblue]{hyperref}
\begin{document}

\title{DEFT-LLM: Disentangled Expert Feature Tuning for Micro-Expression Recognition}

 \author{Ren Zhang, Huilai Li, Chao qi, Guoliang Xu, Tianyu Zhou, Wei wei and Jianqin Yin
 \thanks{Ren Zhang, Huilai Li, Chao qi, Guoliang Xu, Tianyu Zhou, Wei wei and Jianqin Yin are with College of Intelligent Engineering and Automation, Beijing University of Posts and Telecommunications, Beijing, Beijing 100085, China. E-mail: zhangren@bupt.edu.cn, jqyin@bupt.edu.cn}

 }

\markboth{Journal of \LaTeX\ Class Files,~Vol.~14, No.~8, November~2025}%
{Shell \MakeLowercase{\textit{et al.}}: A Sample Article Using IEEEtran.cls for IEEE Journals}


\maketitle

\begin{abstract}
Micro expression recognition (MER) is crucial for inferring genuine emotion. Applying a multimodal large language model (MLLM) to this task enables spatio-temporal analysis of facial motion and provides interpretable descriptions. However, there are still two core challenges: (1) The entanglement of static appearance and dynamic motion cues prevents the model from focusing on subtle motion; (2) Textual labels in existing MER datasets do not fully correspond to underlying facial muscle movements, creating a semantic gap between text supervision and physical motion. To address these issues, we propose DEFT-LLM, which achieves motion semantic alignment by multi-expert disentanglement. We first introduce Uni-MER, a motion-driven instruction dataset designed to align text with local facial motion. Its construction leverages dual constraints from optical flow and Action Unit (AU) labels to ensure spatio-temporal consistency and reasonable correspondence to the movements. We then design an architecture with three experts to decouple facial dynamics into independent and interpretable representations (structure, texture, and motion-semantics). By integrating the instruction-aligned knowledge from Uni-MER into DEFT-LLM, our method injects effective physical priors for micro expressions while also leveraging the cross modal reasoning ability of large language models, thus enabling precise capture of subtle emotional cues. Experiments on multiple challenging MER benchmarks demonstrate state-of-the-art performance, as well as a particular advantage in interpretable modeling of local facial motion.
\end{abstract}


\begin{figure}[ht]
  \centering
  \includegraphics[width=\linewidth]{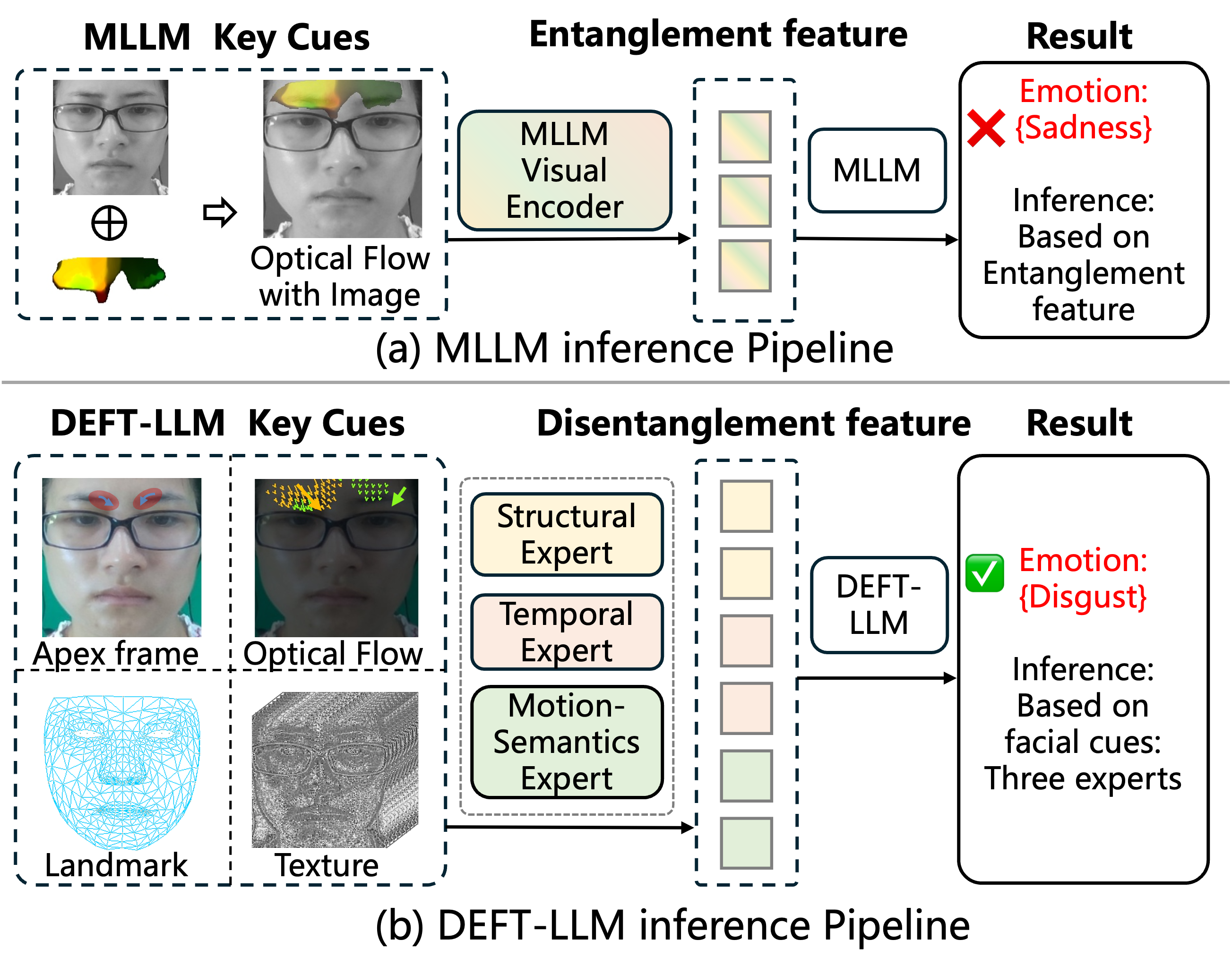}
  \caption{Comparison of emotion reasoning between the MLLM (a), which misinterprets the emotion as {Sadness} due to entanglement feature. In contrast, our DEFT-LLM (b), employing multiple experts to encode distinct key clues, obtains disentangled features. By learning the correspondence between these clues and emotions, it correctly infers the answer as {disgust}.}
  \label{fig:introduction}
\end{figure}

\section{Introduction}
\label{sec:intro}

    Micro-expression recognition (MER) plays a crucial role in uncovering genuine emotions in high-stakes scenarios such as security screening, clinical diagnosis, and negotiation \cite{porter2008reading,yan2013fast}. Unlike conventional facial expressions, micro-expressions occur within 0.2–0.5 seconds, exhibit low intensity, and involve highly localized facial muscle movements \cite{yan2013fast, ekman1969nonverbal}. The central challenge lies in bridging low-level physical dynamic and fine-grained spatiotemporal changes in facial regions \cite{zhao2007dynamic, liong2018less} with high-level semantic understanding that interprets the underlying emotional state \cite{valstar2006fully, happy2014automatic}.
    
    Traditional MER methods  utilize various visual cues, such as RGB frames and optical flow, yet they often lack sufficient semantic reasoning capabilities \cite{zhang2022balance, nguyen2023micron, zhang2022your, fan2023selfme, liu2015main}. While Multimodal Large Language Models (MLLMs) offer a promising direction to enhance this reasoning ability \cite{achiam2023gpt, team2023gemini, lian2024gpt}, they usually rely on a single visual encoder. This architecture tends to produce entangled representations where appearance and motion information are combined \cite{zhang2025mellm}, which can cause the model to overlook the subtle facial dynamics independently conveyed by each cue. Therefore, it is necessary to design a mechanism that extracts expert features from each modality independently, while delegating high-level semantic fusion and reasoning to the LLM.
    
    The second challenge involves the quality of the instruction data used for MER. Prior works \cite{zhang2025mellm, cheng2024emotion} often use LLMs to generate Chain-of-Thought (CoT) descriptions for model fine-tuning. A key issue, however, is that these generated texts are not directly linked to observable facial evidence. This creates a potential mismatch between the textual supervision and the actual visual information in the video frames. More fundamentally, this approach misaligns the learning objective. MER is inherently a discriminative task, aimed at accurately classifying emotions. By training the model to reproduce verbose CoT descriptions, the objective shifts toward a generative task, where the focus becomes mimicking a specific text format. Hence, designing instruction labels that align with the task's discriminative objective is a critical next step.
    
    To address these challenges, we introduce DEFT-LLM. First, to solve the issue of instruction data that lacks connection to real facial movements, we present Uni-MER (Sec. \ref{sec:3.2}), a new dataset where each AU label is linked to observable, region-specific optical flow patterns. This approach bridges the gap between descriptions and actual facial activity. Second, to avoid mixing up different visual signals, we propose the Disentangled Expert Feature Tuning (DEFT) approach (Sec. \ref{sec:3.3}). DEFT-LLM utilizes three separate experts to capture facial structure, dynamic textures, and motion-semantics independently. These features are kept distinct, allowing the model to combine the most important signals rather than treating everything as a single, blended entity. Finally, to keep the training focused on the main goal of recognizing emotions and action units, we use a hybrid training objective (Sec. \ref{sec:3.4}) that emphasizes these classification tasks. Generating captions based on motion remains as an additional training step, helping the model learn about motion patterns without distracting from accurate recognition.
    
    Our main contributions are:
    
    \begin{itemize}
    \item We introduce Uni-MER, the LLM-oriented instruction dataset for micro-expression recognition that grounds symbolic AU labels in region-specific optical flow patterns, enabling motion-aware reasoning with verifiable physical evidence.
    \item We propose DEFT-LLM, a novel disentangled expert architecture that explicitly separates facial structure, dynamic textures, and motion-semantics, and efficiently adapts LLMs to MER through lightweight expert prefix token tuning.
    \item We demonstrate through extensive experiments that DEFT-LLM achieves state-of-the-art performance on challenging CD6ME MER benchmarks , validating the effectiveness of motion-grounded instruction data and disentangled feature learning.
    \end{itemize}

\section{Related Work}
\label{sec:Relate}
\subsection{Traditional Micro-Expression Recognition}
Early micro-expression recognition (MER) research primarily relied on hand-crafted features to capture transient facial changes. LBP-based variants such as LBP-TOP \cite{zhao2007dynamic} demonstrated effectiveness in low-data regimes by extracting spatiotemporal texture descriptors. However, these methods depend heavily on texture information, making them highly susceptible to illumination variations and identity-specific facial appearance patterns.

Recognizing this limitation, researchers shifted their focus toward motion representations, with optical flow emerging as a key signal \cite{liong2018less, liu2015main}. By directly modeling pixel-level inter-frame displacements, optical flow more robustly captures the subtle dynamics of facial muscle movements. To further enhance the signal-to-noise ratio and focus computational attention on relevant motion, researchers began incorporating facial structural priors. In particular, facial landmarks have been used to define regions of interest (ROIs) \cite{zheng2023poster, mao2025poster++}. Aligning optical flow fields or image patches within these ROIs \cite{liu2019neural, zhang2022balance, zhang2022your} enables models to suppress global noise such as head motion and concentrate resources on local muscle groups associated with specific Action Units (AUs) \cite{kumar2022three, zhang2025facial}.

Traditional MER methods achieved substantial progress through explicit spatiotemporal encoding, effectively shifting the task from global texture recognition to structure-guided local motion analysis. Yet a fundamental limitation persists: these methods lack the capacity for high-level semantic reasoning. Operating within a feature-extractor-plus-classifier paradigm, they can detect AU-related motion but cannot reason about it in complex contexts or generate interpretable descriptions. This gap in semantic understanding has motivated recent interest in leveraging Large Language Model.

\subsection{MLLMs for Micro-Expression Recognition}
Recent studies have extended multimodal large language models (MLLMs) to micro-expression recognition (MER). MELLM \cite{zhang2025mellm} represents an early effort, combining optical flow to capture facial motion with appearance features extracted from the original video. These signals are jointly encoded and projected into the model, aiming to highlight dynamic regions relevant to expression analysis. However, the integration of motion and appearance in a single pathway leads to strong feature coupling, which impairs the model’s capacity to disentangle subtle motion cues from static facial textures. As a result, the approach struggles to generalize to variations in micro-expression patterns.

Alternatively, MER-CLIP \cite{liu2025mer} formulates MER as a cross-modal alignment task by translating Action Unit (AU) labels into descriptive text and then associating these with visual features through a CLIP-based framework. While this improves semantic interpretability and facilitates the connection between language and movement, it can introduce semantic drift: the learned associations are not necessarily grounded in measurable physical motion. Moreover, the typical evaluation protocol, such as Leave-One-Subject-Out (LOSO), limits the assessment of model generalization across datasets and raises concerns about robustness in practical applications \cite{varanka2023data}.

Collectively, these recent models point to two major gaps motivating our research. First, when models fail to disentangle motion from appearance, they risk confounding critical motion cues with static background or texture information, leading to performance degradation and bias. Second, when the model’s reasoning relies on abstract textual descriptions (such as LLM-generated CoT or symbolic labels) that lack a verifiable link to the underlying physical motion, its interpretations may become unreliable and difficult to explain. Addressing these gaps in disentanglement and physical grounding is essential for building more robust and interpretable MER systems.

\section{Methodology}
\label{sec:method}
Our method systematically addresses the feature entanglement and semantic drift challenges in the MER task. The approach comprises four interconnected components: first, the task structure design (Sec. \ref{sec:3.1}); second, the motion-anchored instruction dataset Uni-MER (Sec. \ref{sec:3.2}); third, the decoupled expert architecture DEFT-LLM (Sec. \ref{sec:3.3}); and fourth, the hybrid training strategy for objective alignment (Sec. \ref{sec:3.4}).

\subsection{Task Definition}
\label{sec:3.1}

Unlike traditional Micro-Expression Recognition (MER) paradigms that frame the problem as a simple classification task mapping a video to a discrete emotion, we are inspired by MELLM~\cite{zhang2025mellm} to propose an extended task formulation. Our objective is to foster a comprehensive and interpretable understanding of subtle facial behaviors, moving beyond mere categorization.

Formally, given an input micro-expression video clip $V$, our goal is to train a model $\mathcal{M}$ that maps $V$ to a structured, multi-component output $Y$. This output provides a holistic analysis of the facial behavior observed in the video:
\begin{equation}
\label{eq:task_definition}
\hat{y} = \mathcal{M}(V) = (\hat{\mathcal{C}}, \hat{E}, \hat{\mathcal{R}})
\end{equation}
The three components of the output tuple are defined as follows:

\begin{itemize}
    \item \textbf{Category ($\hat{\mathcal{C}}$):} This represents the model's final, high-level semantic prediction. Depending on the specific task, the category $\hat{\mathcal{C}}$ can take one of two forms:
    \begin{itemize}
        \item For Action Unit (AU) recognition, $\hat{\mathcal{C}}$ is a set of one or more activated AU labels, $\hat{\mathcal{C}} = \{\hat{a_1}, \hat{a_2}, \dots\} \subset \hat{\mathcal{A}}$, where $\mathcal{A}$ is the predicted universal set of AUs.
        \item For emotion recognition, $\hat{\mathcal{C}}$ is a single emotion label, $\hat{\mathcal{C}} = \hat{e} \in \hat{\mathcal{O}}$, where $\hat{\mathcal{O}}$ is the set of predicted emotion classes.
    \end{itemize}

    \item \textbf{Evidence ($\hat{E}$):} This component provides the low-level, physical evidence that quantitatively supports the model's prediction. It is a set of region-based motion descriptors, $\hat{E} = \{\hat{E}_i\}_{i=1}^N$, where $N$ is the number of predefined facial regions. This component compels the model to not only make a prediction but also to ground its decision in specific, measurable facial dynamics.

    \item \textbf{Rationale ($\hat{\mathcal{R}}$):} This is a natural language text that logically bridges the motion evidence $\hat{E}$ with the category prediction $\hat{\mathcal{C}}$. The rationale $\hat{\mathcal{R}}$ must articulate the reasoning process, connecting low-level visual signals to high-level semantic concepts, thereby forming a coherent analysis, reasoning, and conclusion chain.
\end{itemize}

By mandating a structured output of $(\hat{\mathcal{C}}, \hat{E}, \hat{\mathcal{R}}     )$, our task paradigm imposes a strong inductive bias: any high-level prediction ($\hat{\mathcal{C}}$) must be grounded in quantifiable, low-level visual evidence ($\hat{E}$) and justified by an intelligible line of reasoning ($\hat{\mathcal{R}}$). This formulation not only enhances the transparency and trustworthiness of the model's decision-making process but also leverages fine-grained supervision to foster a more robust perception and deeper understanding of subtle facial dynamics.
\begin{figure*}[ht]
  \centering
  \includegraphics[width=0.95\linewidth]{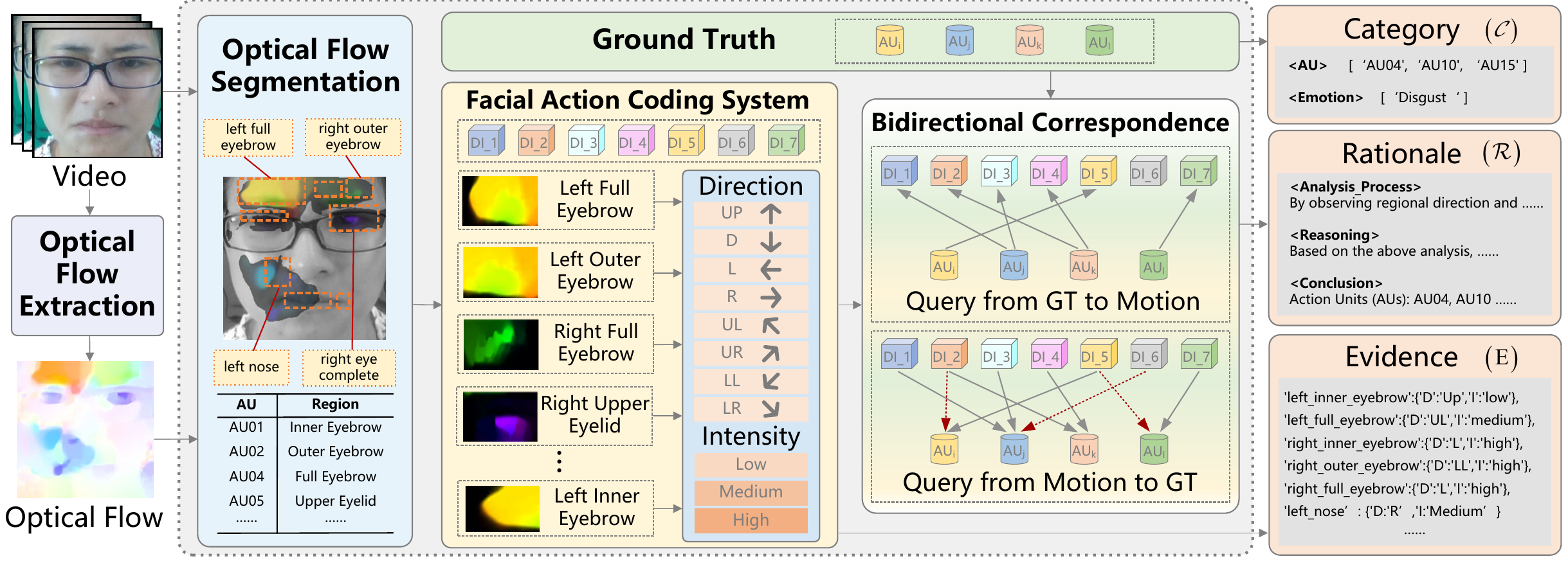}
  \caption{The Uni-MER pipeline: Video is processed into quantified, region-specific Motion Evidence ($\mathbf{E}$). A bidirectional correspondence module (center) verifies this evidence against Ground Truth AUs ($\mathcal{A}$) to generate a grounded Rationale ($\mathcal{R}$), yielding the final structured data triple ($\mathcal{C}, \mathbf{E}, \mathcal{R}$).}
  \label{fig:uni-mer}
\end{figure*}

\subsection{Uni-MER Dataset Construction}
\label{sec:3.2}
To address the limitations of prior MER datasets, where symbolic AU labels are often dissociated from measurable facial dynamics, we introduce the Uni-MER dataset, designed to ground AU annotation in quantifiable motion evidence, as outlined in Algorithm \ref{alg:uni-mer-construction} and Figure \ref{fig:uni-mer}.

The Uni-MER video collection ($\mathcal{V}$) comprises 8,041 samples, processed via a fully automated pipeline from nearly all public micro-expression datasets: $\mathrm{CAS(ME)^2}$ \cite{Qu2017CASME2}, SAMM \cite{Davison2016SAMM}, 4DME \cite{Li2023FourDME}, $\mathrm{CAS(ME)^3}$ \cite{Li2022CASME3}, $\mathrm{CASME \ II}$ \cite{Yan2014CASMEII}, MMEW \cite{Zhao2022SurveyMMEW}, DFME \cite{Zhao2023DFME}, and CASME \cite{Yan2013CASME}. For each video, facial regions ($R_i$) associated with specified AUs are segmented (see Appendix \ref{sec:appendix_a}), and region-wise optical flow features are extracted. 

To suppress global head motion and emphasize local facial dynamics, we apply motion compensation (see Appendix \ref{sec:appendix_b}), obtaining compensated flow vectors $\mathbf{F}_{\text{comp}}(\mathbf{p})$ for pixel $\mathbf{p}$ in region $R_i$.We then quantify local motion by computing two features per region as the motion evidence $E_i$: the dominant direction $\theta_i$ and peak intensity $m_i$, as defined in Eq.\ref{eq:1}:
\begin{equation}
\begin{aligned}
\label{eq:1}
    E_i = (\theta_i, m_i) = \left(\text{atan2}(\overline{F_y}, \overline{F_x}), \frac{1}{|P_{\text{top}}|} \sum_{\mathbf{p} \in P_{\text{top}}} \|\mathbf{F}_{\text{comp}}(\mathbf{p})\|\right)
\end{aligned}
\end{equation}

Where $\theta_i$ is the dominant motion direction calculated by the average motion vector $(\overline{F_x}, \overline{F_y})$ within the region, and is quantized into eight base directions, $m_i $is the average intensity of the K\% ($P_{\text {top}}$) pixels with the strongest motion amplitude in the region $R_i$, discretized into ordinal levels (weak/medium/strong). 

This process converts continuous flow signals into a complete set of standardized descriptors ($\mathbf{E} = {E_i}$).The core of our dataset is the generation of a high-quality rationale ($\mathcal{R}$) that bridges the Ground Truth (GT) AU labels ($\mathcal{A}$) and the quantified motion evidence ($\mathbf{E}$).We generate $\mathcal{R}$ using a dual-verification rationale generation process :
\begin{itemize}

\item \textbf{GT-driven Description (Forward Verification):} We iterate through each GT AU labels ($a \in \mathcal{A}$) and query its corresponding motion evidence ($E_i \in \mathbf{E}$). The rationale ($\mathcal{R}$) records signals (direction and intensity) in the region corresponding to $a$ that are consistent with the expectation for $a$, even if they are very subtle.
\item \textbf{Motion-driven Verification (Backward Verification):} We iterate through all motion evidence with significant magnitude ($E_i$, at a 'strong' level) and map it back to the AU labels ($\mathcal{A}$). If significant motion is detected in a region without any corresponding $a$, or if the motion contradicts the labeled AU, the rationale ($\mathcal{R}$) explicitly flags it as noisy motion.
\end{itemize}

Each sample in Uni-MER is formatted as a structured instruction triple $\mathcal{D}$: the original Ground Truth category labels $\mathcal{C}$ (with AU label $\mathcal{A}$ and emotion label $e$), the complete quantitative motion evidence ($\mathbf{E}$), and the generated rationale text ($\mathcal{R}$). We constrain the explanation ($\mathcal{R}$) to be logically consistent with both the GT labels and the extracted evidence, compelling models to learn the complex correspondence and potential discrepancies between facial dynamics and symbolic tags. Compared to prior datasets that rely on subjective or generative annotation, Uni-MER prioritizes annotation quality and physical traceability, providing a standardized and motion-grounded resource for data-driven AU recognition and reasoning. Further implementation details are in the Appendix \ref{sec:appendix_c}.

\begin{algorithm}[t]
\caption{Uni-MER: Motion-Grounded Rationale Construction}
\label{alg:uni-mer-construction}
\begin{algorithmic}[1]
\Require $\mathcal{V} = \{V_j\}_{j=1}^{M}$: video sets; $\mathcal{A}$: GT AU labels, $e$: GT Emotion label, $\mathcal{C}$: GT Category set with $\mathcal{A}$ and $e$.
\Ensure $\mathcal{D}$: dataset with $(\mathcal{C}, \mathbf{E}, \mathcal{R})$ triples
\State $\mathcal{D} \leftarrow \varnothing$
\For{each $V_j \in \mathcal{V}$}
    \State $\{R_i\}_{i=1}^{N} \gets \text{ExtractRegions}(V_j)$ \Comment{Appendix \ref{sec:appendix_a}}
    \State $\mathbf{E} \leftarrow \varnothing$
    \For{each $R_i$}
        \State $E_i \gets \text{ComputeMotion}(R_i, V_j)$ \Comment{Via Eq.\eqref{eq:1}}
        \State $\mathbf{E} \leftarrow \mathbf{E} \cup \{E_i\}$
    \EndFor
    \State $\mathcal{R} \leftarrow \varnothing$ \quad \Comment{Initialize rationale}
    
    \Comment{Dual-verification for Rationale Generation}
    \State $\mathcal{R}_{fwd} \leftarrow \text{Describe}_{A \to \mathbf{E}}(\mathcal{A}, \mathbf{E})$ \label{alg:line:fwd}
    \State $\mathcal{R}_{bwd} \leftarrow \text{Verify}_{\mathbf{E} \to \mathcal{A}}(\{E_i | \text{Sig}(E_i)\}, \mathcal{A})$ \label{alg:line:bwd}
    \State $\mathcal{R} \leftarrow \mathcal{R}_{fwd} \oplus \mathcal{R}_{bwd}$ \Comment{Combine rationales}
    \State $\mathcal{D} \leftarrow \mathcal{D} \cup \{(\mathcal{C}, \mathbf{E}, \mathcal{R})\}$
\EndFor
\State \Return $\mathcal{D}$
\end{algorithmic}
\end{algorithm}

\begin{figure*}[ht]
  \centering
  \includegraphics[width=\linewidth]{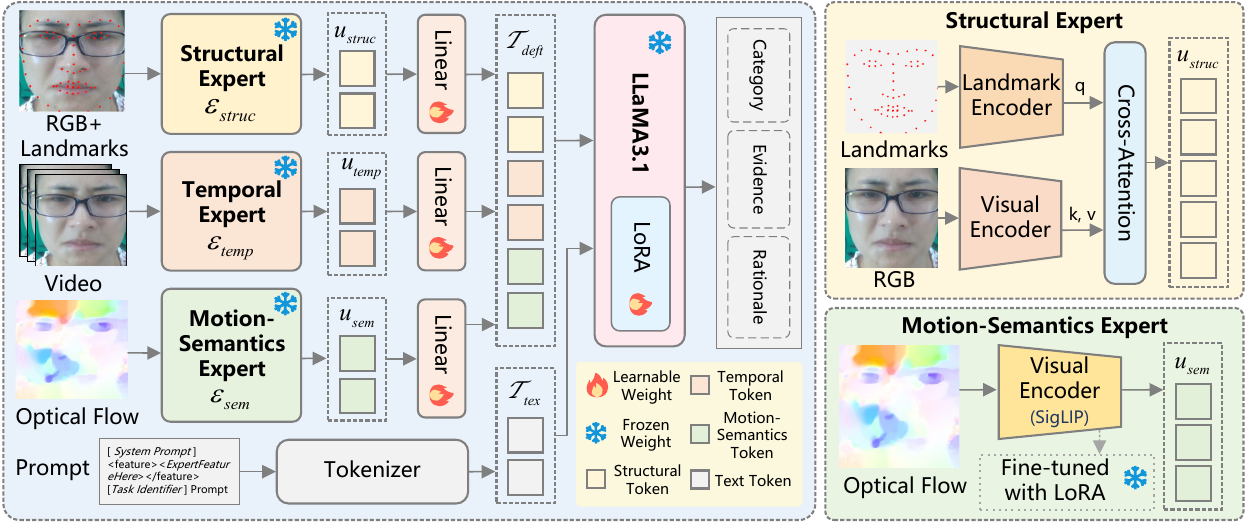}
  \caption{An overview of the DEFT-LLM architecture. Three parallel, frozen expert encoders extract features for structure ($u_{struc}$), temporal dynamics ($u_{temp}$), and motion-semantics ($u_{sem}$) from the different facial cues. These features are projected into expert prefix tokens ($\mathcal{T}_{deft}$) and prepended to the text prompt ($\mathcal{T}_{tex}$). The combined sequence is then processed by a LoRA-tuned LLaMA 3.1 to generate the structured output.}
  \label{fig:deft-llm-arch}
\end{figure*}

\subsection{DEFT-LLM Model}
\label{sec:3.3}
To overcome the feature entanglement common in single-encoder MLLMs, we propose DEFT-LLM, an architecture that disentangles visual representations through three parallel expert encoders specializing in facial structure, temporal dynamics, and motion semantics. As illustrated in Figure \ref{fig:deft-llm-arch}, our model comprises a frozen expert module, $\mathcal{E}_{deft}$, and a LoRA-tuned LLaMA 3.1-8B large language model.

\paragraph{Disentangled Experts.} To capture the subtle, localized facial dynamics overlooked by general-purpose encoders, our expert module, $\mathcal{E}_{deft}$, utilizes three parallel encoders:
\begin{itemize}
\item \textbf{Structural Expert ($\mathcal{E}_{struc}$):} Micro-expressions follow anatomical constraints, different AUs activate specific regions. To model this, we employ facial landmarks from MobileFaceNet \cite{zheng2023poster} as spatial anchors in a cross-attention mechanism. The encoded landmark features ($Q_{lm}$) query static image features ($K_{img}, V_{img}$) from an IR-50 encoder \cite{deng2019arcface} to selectively capture AU-localized appearance, yielding the structure feature $u_{struc}$.
\item \textbf{Temporal Expert ($\mathcal{E}_{temp}$):} Dynamic textures, revealed by appearance changes over time, are effective for delineating emotion categories \cite{zhao2007dynamic}. We use a pre-trained VideoMAE \cite{cheng2024emotion} to extract features from the facial video sequence $V_j$, producing a temporal feature $u_{temp} = \mathcal{E}_{temp}(V_j)$ that captures facial dynamics indicative of emotional states.
\item \textbf{Motion-Semantics Expert ($\mathcal{E}_{sem}$):} While optical flow captures motion, it lacks text semantic alignment. To bridge this gap, we visualize compensated flow vectors $\mathbf{F}_{\text{comp}}$ as an HSV image $\mathcal{F}_{flow}$ and use it to fine-tune a pre-trained SigLIP \cite{zhai2023siglip} visual encoder. This process yields a semantically-aligned motion feature, $u_{sem} = \text{SigLIP}(\mathcal{F}_{flow})$, that links low-level motion patterns to high-level emotional semantics.
\end{itemize}
The features extracted by these three frozen experts[$u_{struc}$, $u_{temp}$, $u_{sem}$] are mapped into the LLM's embedding space via three lightweight, trainable projection layers ($\sigma_{struc}$, $\sigma_{temp}$, $\sigma_{sem}$). This produces a sequence of expert prefix tokens, $\mathcal{T}_{deft}$, which encapsulates the disentangled visual information:
$\mathcal{T}_{deft} = [\sigma_{struc}(u_{struc}), \sigma_{temp}(u_{temp}), \sigma_{sem}(u_{sem})]$
\paragraph{Multimodal Prompting.} To guide the LLM's reasoning, we craft a structured multimodal prompt template inspired by emotion-LLaMA \cite{cheng2024emotion}. This template, detailed in Appendix \ref{sec:appendix_d} and formatted using the LLaMA 3.1 chat structure, integrates system instructions, descriptive captions, and emotion flags to form the textual input, $\mathcal{T}_{tex}$. The complete template structure is as follows:
\begin{center}
[\textit{System Prompt}] \\ \textit{\textless feature\textgreater\textless ExpertFeatureHere\textgreater\textless /feature\textgreater} \\
\textit{[Task Identifier] Prompt}
\end{center}

The final model input is constructed by concatenating the expert prefix tokens with the text prompt. The complete architecture is governed by the following equation:
\begin{equation}
\label{eq:2}
\hat{y} = \Psi_{lora}\left( \mathcal{T}_{deft} \oplus \mathcal{T}_{tex} \right)
\end{equation}
where $\hat{y}$ is the generated structured text, $\Psi_{lora}$ represents the LoRA-tuned LLaMA 3.1, and $\oplus$ denotes sequence concatenation. The LLaMA model's internal cross-attention mechanisms then fuse these expert visual cues with the textual information, enabling nuanced reasoning about the multimodal emotional content. For more details on the network structure, please refer to Appendix \ref{sec:appendix_f}.
\subsection{Hybrid discriminative-generative objective}
\label{sec:3.4}
To resolve the "task-objective mismatch" inherent in generative fine-tuning, where a model might prioritize textual fluency over core discriminative accuracy, we introduce a hybrid discriminative-generative objective. This objective is engineered to forcibly align the model's learning focus with the primary discriminative goals of MER.

Our overall loss function, $\mathcal{L}_{\text{total}}$ , is a weighted sum of generative and discriminative losses:

\begin{equation}
\mathcal{L}_{\text{total}} = \mathcal{L}_{\text{gen}} + w_{\text{emo}} \cdot \mathcal{L}_{\text{emo}} + w_{\text{au}} \cdot \mathcal{L}_{\text{au}}
\end{equation}

where $\mathcal{L}_{\text{gen}}$ is the generative text loss, while  $\mathcal{L}_{\text{emo}} $ represents the discriminative classification losses for Emotion and Action Units (AUs), respectively. To ensure the primacy of the discriminative tasks, we enforce this objective from two complementary perspectives. First, we directly optimize the discriminative losses via a Discriminative Calibration Module (DCM). This module applies two parallel discriminators to the LLM‘s shared last hidden layer for the emotion and AU tasks. During training, we mask the representations of answer text tokens to prevent the discriminators from exploiting textual shortcuts, thereby compelling them to rely solely on visual features. By assigning high weights ($w_{\text{emo}}$  and $w_{\text{au}}$) to these discriminative losses, we guide the shared feature space to preferentially encode information crucial for discerning fine-grained visual discrepancies. Second, we calibrate the generative loss,$ \mathcal{L}{\text{gen}}$, internally. A standard language modeling objective penalizes all tokens uniformly, which can bias optimization toward rationale fluency. To counteract this, we employ a hierarchical weighting mechanism:

\begin{equation}
\mathcal{L}_{\text{gen}} = w_{\text{label}} \cdot\mathcal{L}_{\text{label}} + w_{\text{evi}} \cdot \mathcal{L}_{\text{evi}} + w_{\text{rat}} \cdot \mathcal{L}_{\text{rat}}
\end{equation}

Here, tokens corresponding to the core discriminative labels ($\mathcal{L}_{\text{label}}$) receive a significantly higher weight than those for evidence and rationale. This strategy forces the model to concentrate its optimization on predictive accuracy, treating the generation of interpretable rationales as an auxiliary process that serves the primary task. Through this dual-calibration mechanism, we effectively align the model's generative capabilities with its core perceptual mission.

\begin{table*}[ht]
\caption{Comparison Of The Results Of Different Methods On The Cross-Dataset Protocol.}
\label{tab:All result}
\centering
\resizebox{2\columnwidth}{!}{%
\begin{tabular}{c|ccccccccccccc}
\hline
\textbf{Method} & \textbf{AU1}   & \textbf{AU2}  & \textbf{AU4}  & \textbf{AU5}   & \textbf{AU6}  & \textbf{AU7}  & \textbf{AU9}   & \textbf{AU10}  & \textbf{AU12} & \textbf{AU14} & \textbf{AU15}  & \textbf{AU17}  & \textbf{AVERAGE} \\ \hline

CONSTANT                & 26             & 24.2          & 51.7          & 12.4           & 5.7           & 22.1          & 10.9           & 7.1            & 15.9          & 24            & 5              & 8.1            & 17.8    \\
LBP - TOP \cite{zhao2007dynamic}      & 41.6           & 36.7          & 62            & 0              & 0             & 0             & 1.7            & 0              & 0             & 3.5           & 0              & 0              & 12.1    \\

MDMO \cite{liu2015main}           & 59.4           & 57            & 72.2          & 16.7           & 2.3           & 18.1          & 10.9           & 0              & 27.4          & 25.5          & 13             & 27.8           & 27.5    \\
SVM + OF                & 71             & 65.9          & 84.2          & 13.6           & 10.6 & 44.4          & 17.2           & 10.3           & 26.7          & 39.1          & 18.2           & 32.7           & 36.2    \\

GA - ME \cite{jin2020ga}        & 71.1           & 67.6          & 85.7          & 1.1            & 0             & 26.2          & 7.8            & 2.1            & 12.8          & 15.4          & 0              & 34.7           & 27      \\
OFF-APEXNET \cite{gan2019off}  & 74.9           & 70.2          & 86.3          & 13.5           & 3.3           & 44.5          & \textbf{36.6}           & 18.3           & 32            & 37.7          & 19             & 38.2           & 39.5    \\

STSTNET \cite{liong2019shallow}        & 74.2           & 68.8          & 85.7          & 3.6            & 6.4           & 29            & 18.6           & 9              & 16.7          & 29            & 11.8           & 28.3           & 31.8    \\
RCN-A \cite{xia2020revealing}        & 74.8           & 71            & 85.3          & 4.7            & 0             & 24.1          & 15.9           & 0              & 21.1          & 23.7          & 0              & 24             & 28.7    \\

NMER \cite{liu2019neural}           & 19.1           & 19.2          & 43.3          & 9.3            & 6.3           & 9.1           & 12.5           & 4.8            & 18.1          & 22.6          & 3.3            & 6.4            & 14.5    \\
SSSNET \cite{varanka2021micro}         & 74.6           & 72.1          & \textbf{87.5} & 13.6           & 5.3           & 48.6 & 20.3           & 19.3           & 36.2          & 40.8 & 22.7           & 44.7           & 40.5    \\
RESNET10 \cite{he2016deep}       & 68.7           & 65.2          & 83.8          & 8.3            & 6.4           & 39.6          & 11             & 6.8            & 31            & 29.6          & 8.8            & 39.6           & 33.2    \\
RESNET18 \cite{he2016deep}        & \textbf{76.2}           & 72.1 & 87.1          & 12.6           & 5.5           & 38.2          & 8.6            & 18.4           & 37.3          & 33.7          & 12.4           & 44.7           & 37.2    \\

RESNET34 \cite{he2016deep}        & 72.1           & 71.7          & 85.8          & 11.2           & 5.4           & 38.7          & 16             & 13.5           & 36.5          & 39.9          & 14.8           & 39.2           & 37.1    \\
SCA \cite{li2021micro}            & 42.3           & 42.8          & 56.2          & 14.8           & 1.2           & 23.4          & 13.4           & 3              & 21.8          & 37.1          & 4              & 22.9           & 23.6    \\

LED \cite{varanka2023learnable}            & 52.7           & 45.7          & 63.7          & 7.9            & 0.7           & 19.3          & 13.6           & 8.5            & 26.5          & 36.7          & \textbf{33}             & 31.7           & 28.3    \\
RNET18(2 + 1)D \cite{tran2018closer} & 54.2           & 49.4          & 72.7          & 26.9           & 5.2           & 20.4          & 11.7           & 12.4           & 23.3          & 25.5          & 9.7            & 41.9           & 29.5    \\

\textbf{OURS-unseen}             & 70.68 & \textbf{72.73}         & 84.53         & \textbf{31.63} & \textbf{14.41}         & \textbf{53.73}         & 11.57 & \textbf{29.03} & \textbf{56.24} & \textbf{58.70}         & 12.12& \textbf{47.06} &   \textbf{45.20}    \\ \hline
OURS-LODO       & 86.95          & 88.32         & 89.42         & 65.00          & 46.15         & 75.25         & 12.64           & 55.26          & 80.32         & 80.72         & 35.13          & 54.79          & 64.16   \\ \hline
\end{tabular}
}
\end{table*}

\section{Experiments}
\subsection{Experimental Setup} 
\label{sec:4.1}
We fine-tune the Llama-3.1-8B-Instruct model using LoRA (rank $r=64$) applied to all its attention and feed-forward layers. The model is trained for 40 epochs with a batch size of 1 using the AdamW optimizer and a cosine learning rate scheduler (peak LR: 1e-4). Our hybrid objective's loss weights are set to $w_{au/emo}=1.2$, $w_{label}=2.0$, $w_{evi}=0.6$, and $w_{rat}=0.3$ to prioritize discriminative accuracy. For effective discriminative supervision, we exclusively use the hidden state representations corresponding to the expert prefix tokens.We train on 4*RTX3090 GPUs for 40,000 steps with pipeline parallel module, which takes around 20 hours. All other hyperparameters are detailed in Appendix \ref{sec:appendix_e}.
\subsection{Datasets}
\label{sec:4.2}
We constructed the instruction-following Uni-MER dataset by annotating all samples from our source dataset. The annotation process adhered to the strategy detailed in Sec. \ref{sec:3.2}, and all annotation details were systematically refined.

During this process, we focused on AU and Emotion attributes:
\begin{itemize}
    \item Action Units (AU): We adopted the standard from \cite{varanka2023data}, retaining only the 12 core AUs specified therein (1, 2, 4, 5, 6, 7, 9, 10, 12, 14, 15, 17).
    \item Emotions: We aligned all emotion categories with the DFME \cite{Zhao2023DFME} taxonomy. This mapping covers eight basic emotions (happy, disgust, contempt, surprise, fear, anger, and sad). Any categories not present in DFME were mapped to an {other} class.
\end{itemize}

For evaluation, AU detection accuracy is benchmarked on the six datasets (SAMM \cite{Davison2016SAMM}, 4DME \cite{Li2023FourDME}, $\mathrm{CAS(ME)^3}$ \cite{Li2022CASME3}, $\mathrm{CASME \ II}$ \cite{Yan2014CASMEII}, MMEW \cite{Zhao2022SurveyMMEW}, and CASME \cite{Yan2013CASME}) mentioned in \cite{varanka2023data}. Emotion recognition accuracy is evaluated on the DFME TestA (474 samples) and TestB (299 samples) sets. 

To further evaluate the model's cross-dataset generalization, we designed specific protocols for AU detection and emotion recognition. For AU detection, we employed two settings: (1) a Leave-One-Dataset-Out (LODO) cross-validation following \cite{varanka2023data}, and (2) a stricter setting where all 6 benchmark datasets were strictly excluded from training, using only our remaining around 5,200 samples for training and testing directly on these 6 unseen datasets. For emotion recognition, the evaluation followed a standard cross-domain protocol, where the model is trained on training set, which contains no samples from the target dataset, and evaluated directly on the unseen test sets. Further details on dataset composition are available in Appendix \ref{sec:appendix_c} and \ref{sec:appendix_e}.

\begin{table}
\centering
\caption{Emotion recognition results on the TestA and TestB sets. Best results are in bold.}
\label{tab:mer_results}
\begin{tabular}{l c c c c}
\toprule
MER Methods & Test Set & UF1 & UAR & ACC \\
\midrule
FearRef \cite{zhou2022feature}      & & 0.3410          & 0.3686          & 0.5084 \\
Wang et al. \cite{zhao2024dynamic}  &                        & 0.4067          & 0.4074          & 0.4641          \\
He et al. \cite{zhao2024dynamic}    &     {TestA}                    & 0.4123 & 0.421  & 0.4873          \\
MELLM \cite{zhang2025mellm}      &                        & 0.3578          & 0.3731          & 0.4641          \\
\textbf{DEFT-LLM}       &                        & \textbf{0.4372}          & \textbf{0.4213}          & \textbf{0.5126}          \\
\midrule
FearRef \cite{zhou2022feature}      &  & 0.2875          & 0.3228          & 0.3645          \\
Wang et al. \cite{zhao2024dynamic}  &                        & 0.3534          & 0.3661          & 0.3813          \\
He et al. \cite{zhao2024dynamic}    &     {TestB}                   & 0.4016 & 0.4008 & 0.4147 \\
MELLM \cite{zhang2025mellm}      &                        & 0.3162          & 0.3424          & 0.3712          \\
\textbf{DEFT-LLM}       &                        & \textbf{0.4281}          & \textbf{0.4224}          & \textbf{0.4347}          \\
\bottomrule
\end{tabular}
\end{table}

\subsection{Evaluation Protocol and Metrics}
For evaluation, we parse the predicted emotions and AU categories from the "label" section of the generated text and compare them against the ground truth labels.Our quantitative metrics follow the protocol in \cite{Zhao2023DFME}. For emotion recognition, we report Accuracy (ACC), Unweighted F1 score (UF1), and Unweighted Average Recall (UAR). For AU detection, we report the UF1 score per AU category. Given the prevalent class imbalance in ME datasets, UF1 and UAR serve as crucial metrics for a balanced cross-category performance assessment.


\subsection{Comparison with State-of-the-Art Methods}
\paragraph{Comparison with general AU Detection Methods.} The AU detection task provides objective evidence for capturing and interpreting complex and ambiguous emotional states by deconstructing the subtle dynamics of facial muscles. As shown in TABLE \ref{tab:All result}, in the strict hold-out (unseen) evaluation, DEFT-LLM outperforms the SSSNET \cite{varanka2021micro} baseline, achieving the highest scores in 8 out of 12 AU categories. Although our model was trained on more data, the significant domain gap between our training samples and the unseen test sets highlights the superior generalization capability of DEFT-LLM. Under the LODO protocol (more ME samples compare with unseen), our model substantially outperforms current SOTA methods, with the exception of AU9 (nasal wing). We attribute this specific underperformance to two potential factors: (1) Our optical flow normalization, which subtracts the nose-tip flow vector, may inadvertently suppress the motion signals in the nasal wing region. (2) The 68-point facial landmark system from $\mathcal{E}_{struc}$, used for regional awareness, provides sparse coverage of this area, potentially leading to reduced model attention for this specific motion. This limitation will be addressed in future work.

\paragraph{Comparison with general MER Methods.} As shown in TABLE \ref{tab:mllm}, our method surpasses SOTA \cite{zhao2024dynamic} on UF1 and UAR. We attribute this to the DEFT-LLM architecture: its joint AU/emotion modeling, multi-expert pipelines and DCM guidance compel grounded visual reasoning, outperforming traditional classification models prone to overfitting. However, the improvement margin is modest. We hypothesize this stems from label biases in the training data, leading to superficial correlations. Future work will focus on optimizing reasoning logic and incorporating diverse cues to mitigate this.

\begin{table}[htbp]
\centering
\caption{Comparison with other MLLMS.}
\label{tab:mllm}
\resizebox{1\columnwidth}{!}{%
\begin{tabular}{l c l ccc}
\hline
\textbf{Models}      & \textbf{Class} & \textbf{Dataset}        & \textbf{UF1}            & \textbf{UAR}            & \textbf{ACC}            \\ \hline
Qwen2.5-VL-7B        &  & & 0.0782                  & 0.1886                  & 0.1628                  \\
Qwen-VL-Max          &                    &                           & 0.1039                  & 0.1062                  & 0.1008                  \\
Gemini-2.5-Pro       &     {5}               &         {CASME II}                   & {0.2717}      & {0.2633}      & {0.3023}      \\
MELLM         &                    &                           & {0.4849}         & {0.5337}         & {0.6434}         \\ \cline{4-6}
\textbf{DEFT-LLM (ours)}         &                    &                           & \textbf{0.5349}         & \textbf{0.5717}         & \textbf{0.7596}         \\ \hline
Qwen2.5-VL-7B        &  && 0.1079                  & 0.1787                  & 0.2507                  \\
Qwen-VL-Max          &                    &                           & 0.1623                  & 0.1698                  & 0.1905                  \\
Gemini-2.5-Pro       &                  &                          & {0.1767}      & {0.1663}      & {0.2092}      \\
MELLM          &          {6}            &              {CAS(ME)$^3$}                & {0.2908}         & {0.2948}         & {0.4226}         \\ \cline{4-6}
DEFT-LLM(unseen)      &                    &                           & {0.2617}         & {0.3059}         & {0.2637}         \\ 
\textbf{DEFT-LLM (ours)}      &                    &                           & \textbf{0.3169}         & \textbf{0.3323}         & \textbf{0.3233}         \\ \hline
Qwen2.5-VL-7B        & && 0.0794                  & 0.1495                  & 0.2046                  \\
Qwen-VL-Max          &                    &                           & 0.1788                  & 0.1883                  & 0.2025                  \\
Gemini-2.5-Pro       &       {7}              &              {DFME }               & {0.3056}      & {0.3363}      & {0.3481}      \\
MELLM          &                     &                  TestA         & {0.3578}         & {0.3731}         & {0.4641}         \\ \cline{4-6}
\textbf{DEFT-LLM (ours)}        &                    &                           & \textbf{0.4372}         & \textbf{0.4213}         & \textbf{0.5126}         \\ \hline
Qwen2.5-VL-7B                      &                    & & 0.0737                  & 0.1499                  & 0.1605                  \\
Qwen-VL-Max          &                    &                           & 0.1550                  & 0.1677                  & 0.1672                  \\
Gemini-2.5-Pro       &         {7}           &          {DFME }                  & {0.2665}      & {0.2818}      & {0.2843}      \\
MELLM          &                    &              TestB             & {0.3162}         & {0.3424}         & {0.3712}         \\ \cline{4-6}
\textbf{DEFT-LLM (ours)}        &                    &                           & \textbf{0.4281}         & \textbf{0.4224}         & \textbf{0.4347}         \\ \hline
\end{tabular}
}
\end{table}

\paragraph{Comparison with Other MLLMs.} We also benchmarked against general-purpose MLLMs (e.g., Qwen, Gemini \cite{bai2023qwen, team2023gemini}), which, as shown in TABLE \ref{tab:mllm}, consistently fail to capture subtle ME motions and often default to "Neutral". In contrast, DEFT-LLM leverages its multi-expert architecture to introduce crucial domain priors, enabling more efficient use of emotional representations. To further probe the model's generalization limits, we tested a DEFT-LLM-unseen variant, which excluded $\mathrm{CAS(ME)^3}$ MaE data from its training. As expected, this variant showed a significant performance drop on $\mathrm{CAS(ME)^3}$. However, this drop is attributable to the confounding factors of both reduced training data and the domain shift, making it difficult to decouple their respective impacts in this setting.Despite this confounded result, the model's substantial performance gains on the (also unseen) CASME II benchmark provide strong evidence of its robust generalization capabilities. For more quantitative experimental details, please refer to Appendix \ref{sec:appendix_h}.

\subsection{Ablation Studies}
To validate the key architectural and methodological choices of our proposed model, we conducted a series of additional ablation studies. These experiments are designed to quantify the contribution of each expert encoder and the efficacy of our hybrid loss mechanism, thereby justifying the final model design. All experiments were performed on the challenging DFME TES A and DFME TEST datasets.
\paragraph{Efficacy of the Discriminative Calibration Module (DCM).}
To verify the effectiveness of our hybrid generative-discriminative training approach, we compare the performance of our full model against a variant trained using only the generative loss ($\mathcal{L}_{gen}$), without the auxiliary classification losses from the DCM. The results are shown in Table~\ref{tab:ablation_dcm}.

\begin{table}[ht!]
\centering
\caption{Ablation study on the effect of the Discriminative Calibration Module (DCM). The full model incorporates both generative ($\mathcal{L}_{gen}$) and discriminative ($\mathcal{L}_{DCM}$) losses.}
\label{tab:ablation_dcm}
\resizebox{\columnwidth}{!}{%
\begin{tabular}{l|ccc|ccc}
\toprule
\textbf{Loss Components} & \multicolumn{3}{c|}{\textbf{DFME\_TEST\_A}} & \multicolumn{3}{c}{\textbf{DFME\_TEST\_B}} \\
& UF1 & UAR & ACC & UF1 & UAR & ACC \\
\midrule
$\mathcal{L}_{gen}$ only & 41.38 & 41.68 & 47.47 & 40.30 & 41.81 & 42.96 \\
Full Model ($\mathcal{L}_{gen} + \mathcal{L}_{DCM}$) & \textbf{43.72} & \textbf{42.13} & \textbf{51.26} & \textbf{42.81} & \textbf{42.24} & \textbf{43.47} \\
\bottomrule
\end{tabular}%
}
\end{table}

Training the model with a purely generative objective already achieves strong performance, demonstrating the capability of the LLM to learn the task from rationale generation alone. However, the introduction of the DCM provides a clear and consistent performance enhancement, boosting UF1 on Test A from 41.38\% to 43.72\%.

This improvement can be attributed to the regularization effect of the auxiliary discriminative losses ($\mathcal{L}_{au}$ and $\mathcal{L}_{emo}$). While the generative loss trains the model to produce plausible output sequences, the DCM forces the model's internal representations, specifically the pooled hidden states corresponding to the expert features to be linearly separable for the AU and emotion classification tasks. This dual objective encourages the model to learn a more structured, robust, and calibrated feature space. By explicitly constraining the latent representations to be discriminative, the DCM effectively regularizes the generative process, leading to a model that not only generates better rationales but also makes more accurate final predictions. This confirms that our hybrid loss strategy is superior to a purely generative approach for this task. For more Ablation Studies experimental details, please refer to Appendix \ref{sec:appendix_g}.

\section{Conclusion}
This paper proposes DEFT-LLM, a novel architecture for micro-expression recognition that systematically addresses feature entanglement and semantic drift in existing models. The core of DEFT-LLM is its disentangled expert feature tuning mechanism, which processes facial structure, appearance texture, and motion semantics through parallel expert encoders. To ensure the physical grounding of the model's reasoning, we developed the motion-grounded Uni-MER instruction dataset, which directly links Action Unit labels to quantifiable optical flow patterns. Combined with a hybrid discriminative-generative training objective, our approach significantly enhances the model's perceptual accuracy for subtle dynamics. Comprehensive evaluations demonstrate that DEFT-LLM achieves state-of-the-art performance on multiple challenging benchmarks.Future research will be directed towards integrating more modalities, such as acoustic and physiological signals. We aim to design more reasonable interaction mechanisms to achieve a more precise and holistic perception of emotional states.

\bibliographystyle{IEEEtran}
\bibliography{main}

@String(CVPR= {IEEE Conf. Comput. Vis. Pattern Recog.})

@String(ICCV= {Int. Conf. Comput. Vis.})

@String(ICLR = {Int. Conf. Learn. Represent.})

@String(CVPRW= {IEEE Conf. Comput. Vis. Pattern Recog. Worksh.})

@String(CVPR  = {CVPR})

@String(ICCV  = {ICCV})

@String(ICLR  = {ICLR})

@String(CVPRW= {CVPRW})

@article{porter2008reading,
  title={Reading between the lies: Identifying concealed and falsified emotions in universal facial expressions},
  author={Porter, Stephen and Ten Brinke, Leanne},
  journal={Psychological science},
  volume={19},
  number={5},
  pages={508--514},
  year={2008},
  publisher={SAGE Publications Sage CA: Los Angeles, CA}
}

@article{yan2013fast,
  title={How fast are the leaked facial expressions: The duration of micro-expressions},
  author={Yan, Wen-Jing and Wu, Qi and Liang, Jing and Chen, Yu-Hsin and Fu, Xiaolan},
  journal={Journal of nonverbal behavior},
  volume={37},
  number={4},
  pages={217--230},
  year={2013},
  publisher={Springer}
}

@article{ekman1969nonverbal,
  title={Nonverbal leakage and clues to deception},
  author={Ekman, Paul and Friesen, Wallace V},
  journal={Psychiatry},
  volume={32},
  number={1},
  pages={88--106},
  year={1969},
  publisher={Taylor \& Francis}
}

@article{zhao2007dynamic,
  title={Dynamic texture recognition using local binary patterns with an application to facial expressions},
  author={Zhao, Guoying and Pietikainen, Matti},
  journal={IEEE transactions on pattern analysis and machine intelligence},
  volume={29},
  number={6},
  pages={915--928},
  year={2007},
  publisher={IEEE}
}

@inproceedings{valstar2006fully,
  title={Fully automatic facial action unit detection and temporal analysis},
  author={Valstar, Michel and Pantic, Maja},
  booktitle={2006 conference on computer vision and pattern recognition workshop (CVPRW'06)},
  pages={149--149},
  year={2006},
  organization={IEEE}
}

@article{liong2018less,
  title={Less is more: Micro-expression recognition from video using apex frame},
  author={Liong, Sze-Teng and See, John and Wong, KokSheik and Phan, Raphael C-W},
  journal={Signal Processing: Image Communication},
  volume={62},
  pages={82--92},
  year={2018},
  publisher={Elsevier}
}

@article{happy2014automatic,
  title={Automatic facial expression recognition using features of salient facial patches},
  author={Happy, SL and Routray, Aurobinda},
  journal={IEEE transactions on Affective Computing},
  volume={6},
  number={1},
  pages={1--12},
  year={2014},
  publisher={IEEE}
}

@article{achiam2023gpt,
  title={Gpt-4 technical report},
  author={Achiam, Josh and Adler, Steven and Agarwal, Sandhini and Ahmad, Lama and Akkaya, Ilge and Aleman, Florencia Leoni and Almeida, Diogo and Altenschmidt, Janko and Altman, Sam and Anadkat, Shyamal and others},
  journal={arXiv preprint arXiv:2303.08774},
  year={2023}
}

@article{lian2024gpt,
  title={Gpt-4v with emotion: A zero-shot benchmark for generalized emotion recognition},
  author={Lian, Zheng and Sun, Licai and Sun, Haiyang and Chen, Kang and Wen, Zhuofan and Gu, Hao and Liu, Bin and Tao, Jianhua},
  journal={Information Fusion},
  volume={108},
  pages={102367},
  year={2024},
  publisher={Elsevier}
}

@article{team2023gemini,
  title={Gemini: a family of highly capable multimodal models},
  author={Team, Gemini and Anil, Rohan and Borgeaud, Sebastian and Alayrac, Jean-Baptiste and Yu, Jiahui and Soricut, Radu and Schalkwyk, Johan and Dai, Andrew M and Hauth, Anja and Millican, Katie and others},
  journal={arXiv preprint arXiv:2312.11805},
  year={2023}
}

@article{cheng2024emotion,
  title={Emotion-llama: Multimodal emotion recognition and reasoning with instruction tuning},
  author={Cheng, Zebang and Cheng, Zhi-Qi and He, Jun-Yan and Wang, Kai and Lin, Yuxiang and Lian, Zheng and Peng, Xiaojiang and Hauptmann, Alexander},
  journal={Advances in Neural Information Processing Systems},
  volume={37},
  pages={110805--110853},
  year={2024}
}

@article{zhang2025mellm,
  title={MELLM: Exploring LLM-Powered Micro-Expression Understanding Enhanced by Subtle Motion Perception},
  author={Zhang, Zhengye and Zhao, Sirui and Liu, Shifeng and Yin, Shukang and Mao, Xinglong and Xu, Tong and Chen, Enhong},
  journal={arXiv preprint arXiv:2505.07007},
  year={2025}
}

@inproceedings{nguyen2023micron,
  title={Micron-bert: Bert-based facial micro-expression recognition},
  author={Nguyen, Xuan-Bac and Duong, Chi Nhan and Li, Xin and Gauch, Susan and Seo, Han-Seok and Luu, Khoa},
  booktitle={Proceedings of the ieee/cvf conference on computer vision and pattern recognition},
  pages={1482--1492},
  year={2023}
}

@article{zhang2022balance,
  title={To balance: balanced micro-expression recognition},
  author={Zhang, Ren and He, Ning and Wu, Ying and He, Yuzhe and Yan, Kang},
  journal={Multimedia Systems},
  volume={28},
  number={1},
  pages={335--345},
  year={2022},
  publisher={Springer}
}

@article{zhang2025facial,
  title={Facial 3D Regional Structural Motion Representation Using Lightweight Point Cloud Networks for Micro-Expression Recognition},
  author={Zhang, Ren and Yin, Jianqin and Qi, Chao and Dang, Yonghao and Wang, Zehao and Zhang, Zhicheng and Liu, Huaping},
  journal={IEEE Transactions on Affective Computing},
  year={2025},
  publisher={IEEE}
}

@article{zhang2022your,
  title={Your heart rate betrays you: multimodal learning with spatio-temporal fusion networks for micro-expression recognition},
  author={Zhang, Ren and He, Ning and Liu, Shengjie and Wu, Ying and Yan, Kang and He, Yuzhe and Lu, Ke},
  journal={International Journal of Multimedia Information Retrieval},
  volume={11},
  number={4},
  pages={553--566},
  year={2022},
  publisher={Springer}
}

@inproceedings{fan2023selfme,
  title={SelfME: Self-supervised motion learning for micro-expression recognition},
  author={Fan, Xinqi and Chen, Xueli and Jiang, Mingjie and Shahid, Ali Raza and Yan, Hong},
  booktitle={Proceedings of the IEEE/CVF conference on computer vision and pattern recognition},
  pages={13834--13843},
  year={2023}
}

@article{liu2015main,
  title={A main directional mean optical flow feature for spontaneous micro-expression recognition},
  author={Liu, Yong-Jin and Zhang, Jin-Kai and Yan, Wen-Jing and Wang, Su-Jing and Zhao, Guoying and Fu, Xiaolan},
  journal={IEEE Transactions on Affective Computing},
  volume={7},
  number={4},
  pages={299--310},
  year={2015},
  publisher={IEEE}
}

@article{liu2025mer,
  title={MER-CLIP: AU-Guided Vision-Language Alignment for Micro-Expression Recognition},
  author={Liu, Shifeng and Mao, Xinglong and Zhao, Sirui and Li, Peiming and Xu, Tong and Chen, Enhong},
  journal={IEEE Transactions on Affective Computing},
  year={2025},
  publisher={IEEE}
}

@inproceedings{liu2019neural,
  title={A neural micro-expression recognizer},
  author={Liu, Yuchi and Du, Heming and Zheng, Liang and Gedeon, Tom},
  booktitle={2019 14th IEEE international conference on automatic face \& gesture recognition (FG 2019)},
  pages={1--4},
  year={2019},
  organization={IEEE}
}

@article{mao2025poster++,
  title={Poster++: A simpler and stronger facial expression recognition network},
  author={Mao, Jiawei and Xu, Rui and Yin, Xuesong and Chang, Yuanqi and Nie, Binling and Huang, Aibin and Wang, Yigang},
  journal={Pattern Recognition},
  volume={157},
  pages={110951},
  year={2025},
  publisher={Elsevier}
}

@inproceedings{zheng2023poster,
  title={Poster: A pyramid cross-fusion transformer network for facial expression recognition},
  author={Zheng, Ce and Mendieta, Matias and Chen, Chen},
  booktitle={Proceedings of the IEEE/CVF International Conference on Computer Vision},
  pages={3146--3155},
  year={2023}
}

@inproceedings{kumar2022three,
  title={Three stream graph attention network using dynamic patch selection for the classification of micro-expressions},
  author={Kumar, Ankith Jain Rakesh and Bhanu, Bir},
  booktitle={Proceedings of the IEEE/CVF conference on computer vision and pattern recognition},
  pages={2476--2485},
  year={2022}
}

@article{varanka2023data,
  title={Data leakage and evaluation issues in micro-expression analysis},
  author={Varanka, Tuomas and Li, Yante and Peng, Wei and Zhao, Guoying},
  journal={IEEE Transactions on Affective Computing},
  volume={15},
  number={1},
  pages={186--197},
  year={2023},
  publisher={IEEE}
}

@inproceedings{Yan2013CASME,
  author    = {Weijie Yan and Xiaobai Li and Honglei Wang and Guoying Zhao and Xiaolan Fu},
  title     = {CASME database: A dataset of spontaneous micro-expressions collected from neutralized faces},
  booktitle = {2013 10th IEEE International Conference and Workshops on Automatic Face and Gesture Recognition (FG)},
  year      = {2013},
  pages     = {1--7},
  doi       = {10.1109/FG.2013.6553799}
}

@article{Yan2014CASMEII,
  author  = {Wen-Jing Yan and Xiaobai Li and Sheng-Jia Wang and Guoying Zhao and Yong-Jin Liu and Yi-Hua Chen and Xiaolan Fu},
  title   = {CASME II: An Improved Spontaneous Micro-Expression Database and the Baseline Evaluation},
  journal = {PLOS ONE},
  year    = {2014},
  volume  = {9},
  number  = {1},
  pages   = {e86041},
  doi     = {10.1371/journal.pone.0086041}
}

@article{Qu2017CASME2,
  author  = {Feng Qu and Sheng-Jia Wang and Wen-Jing Yan and Hanjie Li and Shuhong Wu and Xiaolan Fu and Guoying Zhao},
  title   = {CAS(ME)\^{}2: A Database for Spontaneous Macro-Expression and Micro-Expression Spotting},
  journal = {IEEE Transactions on Affective Computing},
  year    = {2017},
  volume  = {9},
  number  = {4},
  pages   = {424--436},
  doi     = {10.1109/TAFFC.2016.2516386}
}

@article{Davison2016SAMM,
  author  = {A. K. Davison and C. Lansley and N. Costen and K. Tan and M. H. Yap},
  title   = {SAMM: A Spontaneous Micro-Facial Movement Dataset},
  journal = {IEEE Transactions on Affective Computing},
  year    = {2016},
  volume  = {9},
  number  = {1},
  pages   = {116--129},
  doi     = {10.1109/TAFFC.2016.2573832}
}

@article{Li2023FourDME,
  author  = {Xiaobai Li and Yongqi An and Jukka H{\"a}kkinen and Guoying Zhao},
  title   = {4DME: A Spontaneous 4D Micro-Expression Dataset With Multimodalities},
  journal = {IEEE Transactions on Affective Computing},
  year    = {2023},
  volume  = {14},
  number  = {4},
  pages   = {3031--3047},
  doi     = {10.1109/TAFFC.2022.3182342}
}

@article{Li2022CASME3,
  author  = {Jingting Li and Weijie Yan and Shuang Liu and Yihong Wu and Xiaolan Fu and Guoying Zhao},
  title   = {CAS(ME)\^{}3: A Third Generation Facial Spontaneous Micro-Expression Database with Depth Information and High Ecological Validity},
  journal = {IEEE Transactions on Affective Computing},
  year    = {2022},
  note    = {Project page: casme.psych.ac.cn},
}

@article{Zhao2022SurveyMMEW,
  author  = {Guoying Zhao and Xiaobai Li and Yante Li and Haoyu Chen and Xiaohua Huang},
  title   = {Video-based Facial Micro-Expression Analysis: A Survey of Datasets, Features and Algorithms},
  journal = {Computer Vision and Image Understanding},
  year    = {2022},
  note    = {Introduces the Micro-and-Macro Expression Warehouse (MMEW) dataset},
}

@article{Zhao2023DFME,
  author  = {Sirui Zhao and Huaying Tang and Xinglong Mao and Shifeng Liu and Yiming Zhang and Hao Wang and Tong Xu and Enhong Chen},
  title   = {DFME: A New Benchmark for Dynamic Facial Micro-Expression Recognition},
  journal = {arXiv preprint arXiv:2301.00985},
  year    = {2023},
  doi     = {10.48550/arXiv.2301.00985}
}

@inproceedings{deng2019arcface,
  title={ArcFace: Additive Angular Margin Loss for Deep Face Recognition},
  author={Jiankang Deng and Jia Guo and Niannan Xue and Stefanos Zafeiriou},
  booktitle={Proceedings of the IEEE/CVF Conference on Computer Vision and Pattern Recognition (CVPR)},
  pages={4690--4699},
  year={2019}
}

@inproceedings{zhai2023siglip,
  title={Sigmoid Loss for Language Image Pre-Training},
  author={Zhai, Xiaohua and Beyer, Lucas and Kolesnikov, Alexander and Zhai, X and others},
  booktitle={Proceedings of the IEEE/CVF International Conference on Computer Vision (ICCV)},
  year={2023},
  pages={11963--11975},
  organization={IEEE}
}

@inproceedings{jin2020ga,
  title={GA-APEXNET: Genetic algorithm in apex frame network for micro-expression recognition system},
  author={Jin, Qiu-Shi and Xu, Huang-Chao and Liu, Kun-Hong and Liong, Sze-Teng and Gan, YS and Su, Shu-Wen},
  booktitle={Journal of Physics: Conference Series},
  volume={1544},
  number={1},
  pages={012149},
  year={2020},
  organization={IOP Publishing}
}

@article{gan2019off,
	title={Off-apexnet on micro-expression recognition system},
	author={Gan, YS and Liong, Sze-Teng and Yau, Wei-Chuen and Huang, Yen-Chang and Tan, Lit-Ken},
	journal={Signal Processing: Image Communication},
	volume={74},
	pages={129--139},
	year={2019},
	publisher={Elsevier}
}

@inproceedings{liong2019shallow,
	title={Shallow triple stream three-dimensional {CNN} ({STSTNet}) for micro-expression recognition},
	author={Liong, Sze-Teng and Gan, YS and See, John and Khor, Huai-Qian and Huang, Yen-Chang},
	booktitle={2019 14th IEEE International Conference on Automatic Face \& Gesture Recognition (FG 2019)},
	pages={1--5},
	year={2019},
	organization={IEEE}
}

@article{xia2020revealing,
  title={Revealing the invisible with model and data shrinking for composite-database micro-expression recognition},
  author={Xia, Zhaoqiang and Peng, Wei and Khor, Huai-Qian and Feng, Xiaoyi and Zhao, Guoying},
  journal={IEEE Transactions on Image Processing},
  volume={29},
  pages={8590--8605},
  year={2020},
  publisher={IEEE}
}

@inproceedings{he2016deep,
	title={Deep residual learning for image recognition},
	author={He, Kaiming and Zhang, Xiangyu and Ren, Shaoqing and Sun, Jian},
	booktitle={Proceedings of the IEEE conference on computer vision and pattern recognition},
	pages={770--778},
	year={2016}
}

@inproceedings{varanka2023learnable,
  title={Learnable Eulerian dynamics for micro-expression action unit detection},
  author={Varanka, Tuomas and Peng, Wei and Zhao, Guoying},
  booktitle={Scandinavian Conference on Image Analysis},
  pages={385--400},
  year={2023},
  organization={Springer}
}

@inproceedings{tran2018closer,
  title={A closer look at spatiotemporal convolutions for action recognition},
  author={Tran, Du and Wang, Heng and Torresani, Lorenzo and Ray, Jamie and LeCun, Yann and Paluri, Manohar},
  booktitle={Proceedings of the IEEE conference on Computer Vision and Pattern Recognition},
  pages={6450--6459},
  year={2018}
}

@article{varanka2021micro,
  title={Micro-expression recognition with noisy labels},
  author={Varanka, Tuomas and Peng, Wei and Zhao, Guoying},
  journal={Electronic Imaging},
  volume={33},
  pages={1--8},
  year={2021},
  publisher={Society for Imaging Science and Technology}
}

@article{li2021micro,
  title={Micro-expression action unit detection with spatial and channel attention},
  author={Li, Yante and Huang, Xiaohua and Zhao, Guoying},
  journal={Neurocomputing},
  volume={436},
  pages={221--231},
  year={2021},
  publisher={Elsevier}
}

@article{zhou2022feature,
  title={Feature refinement: An expression-specific feature learning and fusion method for micro-expression recognition},
  author={Zhou, L. and Mao, Q. and Huang, X. and Zhang, F. and Zhang, Z.},
  journal={Pattern Recognition},
  volume={122},
  pages={108275},
  year={2022}
}

@misc{zhao2024dynamic,
  title={Dynamic micro-expression automatic recognition challenge on the fourth chinese conference on affective computing},
  author={Zhao, S. and Tang, H. and Mao, X. and Liu, S.},
  howpublished={\url{https://mea-lab-421.github.io/CCAC-page/}},
  year={2024},
  note={Accessed: 2025-04-26}
}

@article{bai2023qwen,
  title={Qwen-vl: A versatile vision-language model for understanding, localization, text reading, and beyond},
  author={J. Bai and S. Bai and S. Yang and S. Wang and S. Tan and P. Wang and J. Lin and C. Zhou and J. Zhou},
  journal={arXiv preprint arXiv:2308.12966},
  year={2023},
  note={Online; available at \url{https://arxiv.org/abs/2308.12966}}
}

@book{ekman1997facial,
  title={Facial action coding system},
  author={Ekman, Paul and Friesen, Wallace V},
  year={1997},
  publisher={Consulting Psychologists Press}
}

@inproceedings{kartynnik2019real,
  title={Real-time facial surface geometry from monocular video on mobile gpus},
  author={Kartynnik, Yury and Ablavatski, Artsiom and Grishchenko, Ivan and Grundmann, Matthias},
  booktitle={Proceedings of the IEEE/CVF Conference on Computer Vision and Pattern Recognition Workshops},
  pages={0--0},
  year={2019}
}

@inproceedings{loshchilov2019decoupled,
  title={Decoupled Weight Decay Regularization},
  author={Loshchilov, Ilya and Hutter, Frank},
  booktitle={International Conference on Learning Representations (ICLR)},
  year={2019},
  url={https://arxiv.org/abs/1711.05101}
}

@inproceedings{tong2022videomae,
  title={VideoMAE: Masked Autoencoders are Data-Efficient Learners for Self-Supervised Video Pre-Training},
  author={Tong, Zhan and Song, Yibing and Wang, Jue and Wang, Limin},
  booktitle={Advances in Neural Information Processing Systems (NeurIPS)},
  year={2022},
  url={https://arxiv.org/abs/2203.12602}
}

\vfill

\clearpage
\setcounter{page}{1}

\appendices
\section{AU-Associated Facial Region of Interest (ROI) Definition}
\label{sec:appendix_a}

This appendix provides a detailed specification of the facial Regions of Interest (ROIs) associated with specific Action Units (AUs), as referenced in Sec. \ref{sec:3.2} of the main paper. Our methodology is designed to be precise, reproducible, and anatomically grounded, forming the basis for our subsequent motion analysis.

\subsection{ROI Design Principles}
The definition of our 17 ROIs is guided by the following principles:
\begin{itemize}
    \item \textbf{Anatomical Correspondence:} Each ROI is meticulously defined to correspond to the facial muscle regions activated by specific AUs, as described in the Facial Action Coding System (FACS)~\cite{ekman1997facial}.
    \item \textbf{Bilateral Symmetry:} For AUs that can occur asymmetrically, we define distinct left and right ROIs (e.g., for eyebrows and eye regions) to enable the analysis of expression asymmetry.
    \item \textbf{Granularity and Refinement:} For complex facial areas like the eyebrows, we define sub-regions (inner, outer) in addition to the complete region. This allows for a more granular analysis, distinguishing between AUs such as AU1 (Inner Brow Raiser) and AU2 (Outer Brow Raiser).
    \item \textbf{Landmark-Driven Definition:} All ROI boundaries are programmatically determined by a set of facial landmarks, ensuring objectivity and perfect reproducibility.
\end{itemize}

\subsection{Technical Implementation}
\subsubsection{Facial Landmark System}
Our approach leverages a dense 468-point facial landmark model, akin to the one provided by MediaPipe Face Mesh~\cite{kartynnik2019real}. This dense model offers superior detail compared to sparser 68-point models, which is crucial for accurately delineating the subtle contours of ROIs relevant to micro-expressions, particularly around the eyes and eyebrows. Each landmark is represented by a 2D coordinate $(x, y)$.

\subsubsection{ROI Segmentation Algorithm}
To generate a binary mask for each ROI from its corresponding set of landmark points, we employ the Convex Hull algorithm. The process is as follows:
\begin{enumerate}
    \item \textbf{Landmark Extraction:} For a given ROI, we first retrieve the coordinates of its pre-defined set of landmark indices (see Table~\ref{tab:au_roi_mapping}) from the full 468-point mesh.
    \item \textbf{Convex Hull Computation:} We compute the convex hull of these landmark points using the "scipy.spatial.ConvexHull" library. The convex hull provides a tight, enclosing polygon for the points, effectively outlining the region's boundary.
    \item \textbf{Mask Generation:} The vertices of the resulting convex hull polygon are then used to generate a binary mask of the same dimensions as the input image. We use the Python Imaging Library (PIL) to draw and fill the polygon, where pixels inside the ROI are set to 1 (or "True") and 0 (or "False") otherwise.
    \item \textbf{Robustness Handling:} In the rare case that the convex hull computation fails (e.g., due to collinear points), the algorithm gracefully falls back to generating a simple bounding box around the landmark points to ensure robustness.
\end{enumerate}
This method is highly adaptive to the non-rigid and curved nature of facial regions and robust to variations in face shape and orientation.

\begin{figure}[ht!]
    \centering
    \includegraphics[width=0.9\columnwidth]{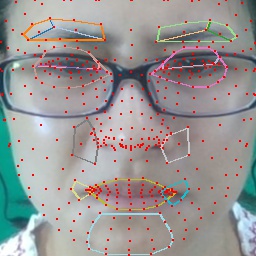} 
    \caption{\textbf{Visualization of the 17 Defined Facial ROIs.} This figure illustrates the 17 regions, which are segmented by applying convex hulls to groups of facial landmarks. The selection of these landmarks is based on prior anatomical knowledge. The final ROIs and their corresponding defining landmarks are shown overlaid on a facial image.}
    \label{fig:roi_visualization}
\end{figure}
\begin{table*}[ht]
\centering
\caption{\textbf{Action Unit (AU) to Region of Interest (ROI) Mapping.} This table details the correspondence between each analyzed AU and its designated facial region(s). The Landmark Indices refer to the 468-point dense facial landmark model.}
\label{tab:au_roi_mapping}
\resizebox{\textwidth}{!}{%
\begin{tabular}{l|l|l|l} 
\toprule
\textbf{AU} & \textbf{FACS Description} & \textbf{Corresponding ROI Name} & \textbf{Landmark Indices (from 468-point model)} \\
\midrule
\hline
\multicolumn{4}{c}{\textbf{Eyebrow Regions}} \\
\hline
AU1 & Inner Brow Raiser & "left\_inner\_eyebrow" & [46, 53, 52, 105, 63, 70] \\
    &                   & "right\_inner\_eyebrow" & [285, 295, 282, 334, 296, 336] \\
\hline
AU2 & Outer Brow Raiser & "left\_outer\_eyebrow" & [52, 65, 55, 107, 66, 105] \\
    &                   & "right\_outer\_eyebrow" & [282, 283, 276, 300, 293, 334] \\
\hline
AU4 & Brow Lowerer & "left\_full\_eyebrow" & [46, 53, 52, 65, 55, 107, 66, 105, 63, 70] \\
    &                & "right\_full\_eyebrow" & [285, 295, 282, 283, 276, 300, 293, 334, 296, 336] \\
\hline
\multicolumn{4}{c}{\textbf{Eye Regions}} \\
\hline
AU5 & Upper Lid Raiser & "left\_upper\_eyelid" & [226, 130, 33, 161, 159, 158, 157, 173, 243, 190, 56, 28, 27, 29, 30, 247] \\
    &                  & "right\_upper\_eyelid" & [463, 398, 384, 385, 386, 387, 388, 466, 263, 467, 260, 259, 257, 258, 286, 414] \\
\hline
AU6 & Cheek Raiser & "left\_lower\_eyelid" & [226, 130, 33, 7, 144, 145, 153, 154, 133, 244, 245, 233, 232, 231, 230, 229, 228, 31] \\
    &              & "right\_lower\_eyelid" & [446, 359, 263, 390, 373, 374, 380, 381, 362, 464, 465, 453, 452, 451, 450, 449, 448, 261] \\
\hline
AU7 & Lid Tightener & "left\_eye\_complete" & Union of "left\_upper\_eyelid" and "left\_lower\_eyelid" indices \\
    &               & "right\_eye\_complete" & Union of "right\_upper\_eyelid" and "right\_lower\_eyelid" indices \\
\hline
\multicolumn{4}{c}{\textbf{Nose, Mouth, and Chin Regions}} \\
\hline
AU9 & Nose Wrinkler & "left\_nose" & [64, 98, 165, 206, 36, 142, 49] \\
    &               & "right\_nose" & [294, 327, 391, 426, 266, 371, 279] \\
\hline
AU10 & Upper Lip Raiser & "mouth" & [61, 40, 39, 37, 0, 267, 269, 270, 291, 321, 405, 314, 17, 84, 181, 91] \\
\hline
AU12 & Lip Corner Puller & "left\_mouth\_corner" & [57, 43, 146, 96, 183, 186] \\
AU14 & Dimpler           & "right\_mouth\_corner" & [287, 273, 375, 325, 407, 410] \\
AU15 & Lip Corner Depressor & & \\
\hline
AU17 & Chin Raiser & "chin" & [17, 18, 83, 182, 194, 32, 140, 176, 148, 152, 377, 400, 369, 262, 418, 406, 313] \\
\bottomrule
\end{tabular}
}
\end{table*}
\subsection{AU-to-ROI Mapping}
Table~\ref{tab:au_roi_mapping} presents the comprehensive mapping between the AUs analyzed in this study and their corresponding facial ROIs. For each AU, we specify the ROI name and the landmark indices used to define it.

\subsubsection{Mapping Rationale}
The relationship between AUs and ROIs follows several patterns:
\begin{itemize}
    \item \textbf{Symmetrical Mapping:} Most AUs (e.g., AU1, AU2, AU4, AU5, AU6) are mapped to bilateral ROIs on the left and right sides of the face.
    \item \textbf{Shared Regions:} Certain AUs activate the same facial area, leading them to share an ROI. For example, AU12 (Lip Corner Puller), AU14 (Dimpler), and AU15 (Lip Corner Depressor) all manifest as movements within the "left/right mouth corner" ROIs. The differentiation between these AUs is achieved by analyzing the distinct characteristics of the motion vector fields (e.g., direction and magnitude) within the shared region, not by the region definition itself.
    \item \textbf{Hierarchical Regions:} The eyebrow and eye areas feature a hierarchical structure. For instance, AU4 (Brow Lowerer) is mapped to the "full eyebrow" ROIs, which encompass the "inner eyebrow" (for AU1) and "outer eyebrow" (for AU2) sub-regions. A similar hierarchy exists for the eye regions concerning AU5, AU6, and AU7.
\end{itemize}

\section{Motion Compensation Technical Details}
\label{sec:appendix_b}

This appendix elaborates on the motion compensation algorithm employed to isolate local facial muscle movements from global head motion, as mentioned in Sec. \ref{sec:3.2} and Algorithm \ref{alg:uni-mer-construction} of the main paper. The objective is to transform the raw optical flow field, which contains a mixture of motions, into a compensated field that primarily reflects muscular articulations pertinent to micro-expressions.

\subsection{Problem Formulation and Objective}
The raw optical flow field, $F_{raw}(p)$, computed between two consecutive frames, represents the displacement vector for each pixel $p$. This field is a superposition of two distinct components:
\begin{itemize}
    \item \textbf{Global Head Motion ($F_{global}$):} Rigid or near-rigid motion caused by the subject's head moving, rotating, or scaling within the frame.
    \item \textbf{Local Facial Muscle Motion ($F_{local}$):} Non-rigid, localized motion resulting from the contraction and relaxation of facial muscles, which constitutes the expression itself.
\end{itemize}
For robust micro-expression analysis, it is imperative to suppress the dominant, often high-magnitude global motion component to accentuate the subtle, low-magnitude local motions. Our objective is thus to compute a compensated flow field, $F_{comp}(p)$, such that $F_{comp}(p) \approx F_{local}(p)$.

\subsection{Reference Point Selection: The Nose Tip}
The cornerstone of our compensation strategy is the subtraction of a reference motion vector that approximates the global head motion. We selected the nose tip as the source for this reference vector, a decision supported by anatomical and kinematic rationale.

\subsubsection{Justification for the Nose Tip}
\begin{itemize}
    \item \textbf{Anatomical Stability:} The nose tip is structurally supported by bone and cartilage and has minimal musculature directly attached to it that would cause significant deformation during most expressions. It resides at the geometric center of the face, making it a stable anchor whose motion is highly correlated with the rigid motion of the skull.
    \item \textbf{Kinematic Inertness:} Unlike the peri-orbital or peri-oral regions, the nose tip exhibits negligible independent motion during the activation of most AUs. Its displacement is almost entirely attributable to the global movement of the head.
    \item \textbf{Detection Reliability:} Modern landmark detectors identify the nose tip with high precision and reliability across various poses and lighting conditions.
\end{itemize}

To further enhance robustness against minor deformations (e.g., nostril flaring in AU9) and landmark jitter, we do not rely on a single point. Instead, we compute the geometric centroid of a stable 4-point cluster on the nose tip (landmark indices: 44, 51, 274, 281 in the 468-point model). This averaging strategy yields a highly stable sub-pixel reference coordinate.

\subsection{Algorithm and Implementation}
The motion compensation process is executed in three sequential steps, detailed below and summarized in Algorithm~\ref{alg:motion_compensation}.

\begin{algorithm}[ht!]
\caption{Motion Compensation Algorithm}
\label{alg:motion_compensation}
\begin{algorithmic}[1]
\Require Raw optical flow field $F_{raw}$ ($H \times W \times 2$)
\Require Facial landmarks $L$ ($468 \times 2$)
\Ensure Compensated optical flow field $F_{comp}$
\Statex
\Function{MotionCompensate}{$F_{raw}, L$}
    \State \Comment{Step 1: Compute nose tip centroid}
    \State $N_{indices} \leftarrow \{44, 51, 274, 281\}$
    \State $N_{points} \leftarrow L[N_{indices}]$
    \State $C_{nose} \leftarrow (\frac{1}{4}\sum_{p \in N_{points}} p.x, \frac{1}{4}\sum_{p \in N_{points}} p.y)$
    \Statex
    \State \Comment{Step 2: Get flow vector at nose tip}
    \State $F_{nose\_tip} \leftarrow \text{BilinearInterpolate}(F_{raw}, C_{nose})$
    \Statex
    \State \Comment{Step 3: Apply compensation}
    \State $F_{comp} \leftarrow F_{raw}$
    \For{each pixel $p$ in $F_{comp}$}
        \If{$\|F_{raw}(p)\| > 10^{-6}$}
            \State $F_{comp}(p) \leftarrow F_{raw}(p) - F_{nose\_tip}$
        \EndIf
    \EndFor
    \State \Return $F_{comp}$
\EndFunction
\end{algorithmic}
\end{algorithm}

\subsubsection{Step 1: Nose Tip Centroid Calculation}
The reference point, $C_{nose} = (c_x, c_y)$, is computed as the arithmetic mean of the coordinates of the four pre-defined nose tip landmarks.
\begin{equation}
C_{nose} = \left( \frac{1}{4}\sum_{i \in N_{indices}} L_{i,x}, \frac{1}{4}\sum_{i \in N_{indices}} L_{i,y} \right)
\label{eq:nose_center}
\end{equation}

\subsubsection{Step 2: Reference Flow Vector Extraction}
Since $C_{nose}$ is a sub-pixel coordinate, we cannot directly index the discrete flow field. We employ bilinear interpolation to sample the flow vector $F_{nose\_tip}$ at $C_{nose}$ with sub-pixel accuracy. This provides a more precise estimate of the motion at the reference point than nearest-neighbor sampling.

\subsubsection{Step 3: Flow Field Compensation}
The final compensated flow, $F_{comp}(p)$, is obtained by subtracting the reference flow vector $F_{nose\_tip}$ from the raw flow vector $F_{raw}(p)$ at every pixel $p$. The formula is:
\begin{equation}
F_{comp}(p) = 
\begin{cases} 
      F_{raw}(p) - F_{nose\_tip} & \text{if } \|F_{raw}(p)\| > \epsilon \\
      F_{raw}(p) & \text{otherwise}
   \end{cases}
\label{eq:compensation}
\end{equation}
where $\epsilon$ is a small threshold (e.g., $10^{-6}$) to prevent applying compensation to static background pixels where the flow is zero. This conditional application avoids introducing minor noise into truly static regions of the image.
\subsubsection{Step 4: Gamma Correction for Contrast Enhancement}
\label{sec:gamma_correction}

Following the motion compensation, we apply a non-linear Gamma correction to the magnitude of the compensated flow vectors, $F_{comp}$. This step is critical for enhancing the signal-to-noise ratio of the motion field before visualization and analysis. The transformation is defined as:
\begin{equation}
M'_{p} = (M_{p})^{\gamma}
\label{eq:gamma}
\end{equation}
where $M_{p} = \|F_{comp}(p)\|$ is the normalized magnitude of the flow vector at pixel $p$, and $M'_{p}$ is the corrected magnitude. The direction of the flow vector remains unchanged.

In our implementation, we use a gamma value of $\boldsymbol{\gamma=2.0}$. The effect of this power-law function is to quadratically scale the motion magnitudes. Consequently, low-magnitude vectors—often corresponding to noise, sensor jitter, or residual global motion artifacts—are significantly suppressed (e.g., a magnitude of 0.2 becomes 0.04). Conversely, high-magnitude vectors, which represent the more salient and intense muscular movements, are preserved or amplified relative to the suppressed noise.

This process acts as a soft threshold, effectively filtering out minor, irrelevant motions and ensuring that the final flow representation predominantly captures the most vigorous parts of the facial expression. 
\subsection{Relation to Prior Work and Visualization}
\begin{figure}[ht!]
    \centering
    \includegraphics[width=\columnwidth]{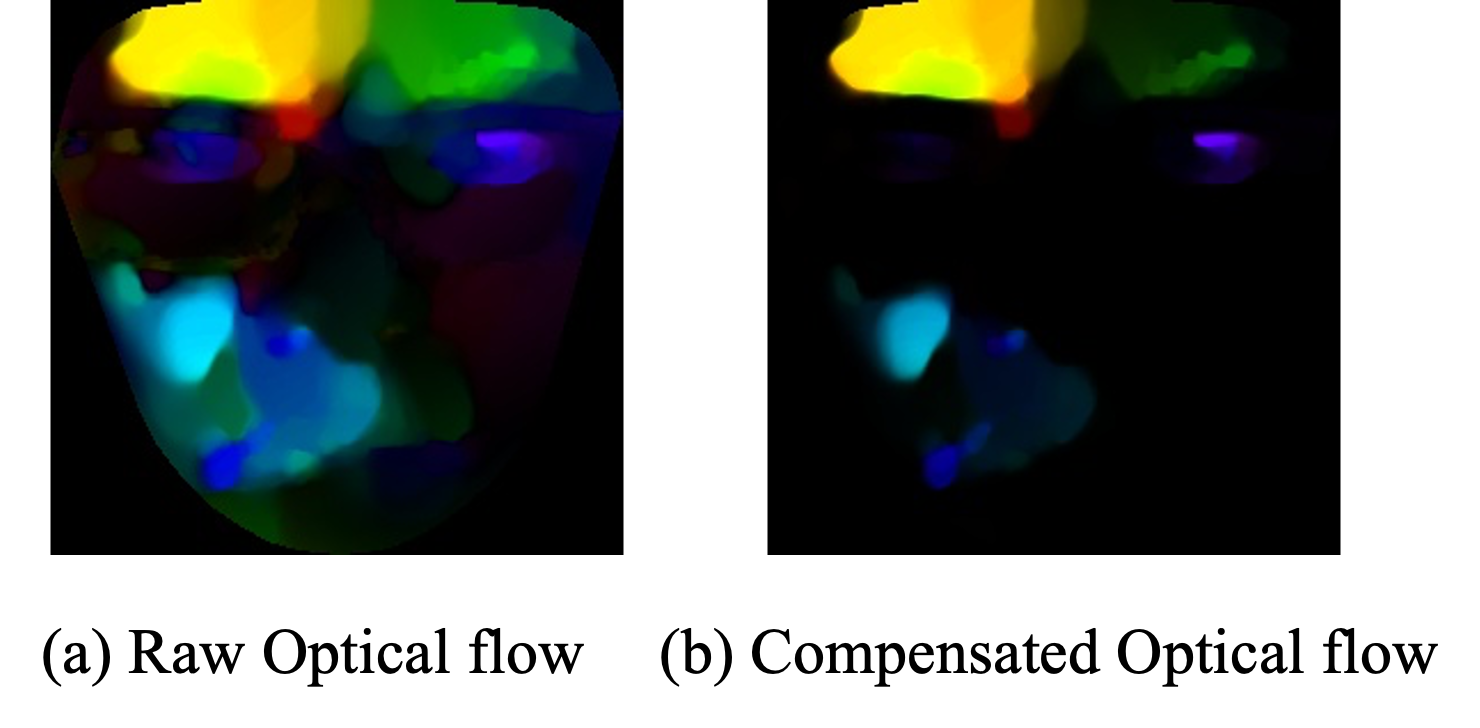}
    \caption{\textbf{Effect of Motion Compensation.} (a) Raw optical flow visualized in HSV, where a uniform color indicates dominant global head motion (e.g., upward-left). (b) Compensated flow for the same frame. The global motion is effectively nullified, revealing the subtle, localized motions corresponding to a facial expression (e.g., contraction around the eyes and mouth corners) as distinct color patterns against a static (black) background.}
    \label{fig:motion_comp_vis}
\end{figure}

Our methodology for extracting and aligning optical flow builds upon the effective strategies demonstrated in recent works~\cite{zhang2022balance, liu2015main}, which have proven successful in highlighting key motion areas. However, we diverge in our subsequent processing. Instead of using a standard color wheel, we map the compensated flow field $F_{comp}$ into the HSV color space for visualization. The hue component directly encodes the motion direction, while the saturation and value components represent the motion magnitude. This choice is deliberate, as it produces a visual representation where motion semantics (direction, speed) are mapped to color properties that can be more effectively interpreted, not only by human observers but also potentially by Large Language Models (LLMs) in multi-modal contexts. The effect of this compensation is visually demonstrated in Figure~\ref{fig:motion_comp_vis}.

\section{Uni-MER Dataset Construction}
\label{sec:appendix_c}

This appendix provides a comprehensive overview of the construction, annotation pipeline, and statistical composition of our Uni-MER dataset. The primary goal of Uni-MER is to amalgamate multiple existing micro-expression datasets into a single, large-scale, and richly annotated resource to facilitate research in automated micro-expression analysis.

The Uni-MER dataset is built upon a curated selection of established public micro-expression databases. This amalgamation is intentionally designed to capture a wide variance in collection environments, subject demographics, temporal resolutions (FPS), and spatial resolutions. Table~\ref{tab:dataset} provides a summary of the key properties of the primary source datasets utilized in this work. This heterogeneity forms a robust foundation for Uni-MER, ensuring that models trained on it are exposed to a rich and varied set of micro-expression exemplars.

\begin{table*}[ht]
\caption{Micro-expression Datasets.}
\label{tab:dataset}
\centering
\begin{tabular}{cccccccc}
\hline
\textbf{Dataset}              & \textbf{Subjects} & \textbf{Micro-expr.} & \textbf{Macro-expr.} & \textbf{Raw Res} & \textbf{Color Mode} & \textbf{FPS} & \textbf{AU Labels} \\ \hline
\multicolumn{1}{c|}{$\mathrm{SAMM}$ \cite{Davison2016SAMM}}     & 32                & 159                  & 343                  & 2040 × 1088      & Gray                & 200          & 27                 \\
\multicolumn{1}{c|}{ $\mathrm{CASME}$ \cite{Yan2013CASME}}    & 35                & 195                  & N/A                  & 640 × 480        & RGB                 & 60           & 21                 \\
\multicolumn{1}{c|}{$\mathrm{CASME II}$ \cite{Yan2014CASMEII}} & 26                & 247                  & N/A                  & 640 × 480        & RGB                 & 200          & 19                 \\
\multicolumn{1}{c|}{$\mathrm{CAS(ME)^2}$ \cite{Qu2017CASME2}} & 22                & 57                   & 300                  & 640 × 480        & RGB                 & 30           & 28                 \\
\multicolumn{1}{c|}{$\mathrm{CAS(ME)^3}$ \cite{Li2022CASME3}} & 247               & 1109                 & 3490                 & $\sim$640 × 480  & RGB                 & 30           & 24                 \\
\multicolumn{1}{c|}{$\mathrm{MMEW} $\cite{Zhao2022SurveyMMEW}}     & 36                & 300                  & 900                  & 1920 × 1080      & RGB                 & 200          & 17                 \\
\multicolumn{1}{c|}{$\mathrm{DFME}$ \cite{Zhao2023DFME}}     & 176               & 1857                 & N/A                  & 1280 × 800       & RGB\&Gray           & 200/300/500  & 22                 \\
\multicolumn{1}{c|}{$\mathrm{4DME}$ \cite{Li2023FourDME}}     & 41                & 267                  & 123                  & 640 × 480        & Gray                & 30/60        & 22                 \\ \hline
\end{tabular}
\end{table*}

\subsection{Dataset Composition}
\label{sec:dataset_composition}

The Uni-MER dataset comprises a total of \textbf{8,041} samples, aggregated from 12 distinct source datasets. This aggregation provides a high degree of diversity in subjects, acquisition conditions, and expression elicitation methods. The distribution of samples from each source is detailed in Table~\ref{tab:dataset_composition}.

The sources can be broadly categorized into three families:
\begin{itemize}
    \item \textbf{CAS(ME) Series:} The most significant contributor, with 4,582 samples (57.0\%) from CASME, CASME II, $\mathrm{CAS(ME)^2}$, $\mathrm{CAS(ME)^3}$, and $\mathrm{CAS(ME)^3\_MaE}$.
    \item \textbf{DFME Series:} A substantial collection of 2,629 samples (32.7\%) from DFME and its associated test sets.
    \item \textbf{Other Datasets:} A diverse set of 830 samples (10.3\%) from MMEW, D4ME, D4ME\_MAE, SMIC, and SAMM, which enriches the dataset's variety.
\end{itemize}

\begin{table}[ht!]
\centering
\caption{\textbf{Source Distribution of Samples in the Uni-MER Dataset.} The dataset aggregates 8,041 samples from 12 distinct sources, enhancing diversity and scale.}
\label{tab:dataset_composition}
\resizebox{0.45\textwidth}{!}{%
\begin{tabular}{l|r|r}
\toprule
\textbf{Source Dataset} & \textbf{Sample Count} & \textbf{Percentage (\%)} \\
\midrule
$\mathrm{CAS(ME)^3\_MaE}$    & 2,945 & 36.62\% \\
DFME            & 1,856 & 23.08\% \\
$\mathrm{CAS(ME)^3}$          & 840   & 10.45\% \\
DFME\_TEST\_A     & 474   & 5.89\%  \\
$\mathrm{CAS(ME)^2}$         & 354   & 4.41\%  \\
DFME\_TEST\_B     & 299   & 3.72\%  \\
MMEW            & 292   & 3.63\%  \\
D4ME            & 265   & 3.31\%  \\
CASME II         & 255   & 3.17\%  \\
CASME        & 188   & 2.33\%  \\
SAMM            & 158   & 1.96\%  \\
D4ME\_MAE       & 115   & 1.43\%  \\
\midrule
\textbf{Total}  & \textbf{8,041} & \textbf{100.00\%} \\
\bottomrule
\end{tabular}%
}
\end{table}

\subsection{Annotation and Rationale Generation Pipeline}
To ensure the quality and utility of our dataset, we developed a systematic pipeline for annotation verification and automated rationale generation.

\subsubsection{Dual-Verification Mechanism}
We implemented a dual-verification mechanism to cross-validate the ground-truth Action Unit (AU) labels with the computed optical flow data for each sample. This process consists of two stages:
\begin{enumerate}
    \item \textbf{Forward Verification:} For each AU specified in the ground-truth label, we analyze the motion within its anatomically corresponding facial Region of Interest (ROI). We quantify the magnitude and direction of the optical flow in this region. If the observed motion, even if subtle, is directionally consistent with the muscle movement defined by the Facial Action Coding System (FACS)~\cite{ekman1997facial} for that AU (e.g., upward motion for AU1/AU2), it is considered a positive verification. This finding is explicitly recorded in the generated rationale ($\mathcal{R}$).
    
    \item \textbf{Reverse Verification and Anomaly Identification:} We systematically scan all predefined facial ROIs for significant motion. If a region exhibits motion above a set intensity threshold but is \textit{not} associated with any of the ground-truth AUs for that sample, this discrepancy is noted. This step serves to identify potential annotation errors or confounding motions. These anomalies are documented in the rationale with a plausible explanation based on the region's context. For instance, significant motion in the peri-orbital region without corresponding AU labels is often attributed to eye blinks, whereas unexplained motion in other regions is flagged as potential noise from illumination changes or other artifacts.
\end{enumerate}

\subsubsection{FACS Correlation and Motion Significance}
The verification process relies on quantitative criteria to correlate optical flow with AU activations.

\paragraph{AU-to-Region Correspondence.} Each AU is mapped to one or more facial ROIs defined by a set of 468 facial landmarks, as detailed in Appendix A. The boundary of each ROI is determined by computing the convex hull of its constituent landmarks, ensuring a tight and anatomically relevant mask.

\paragraph{Motion Intensity Thresholds.} To objectively classify the intensity of facial muscle movement, we defined four levels based on the average magnitude of the top 10\% ($K$)most intense flow vectors within an ROI. These thresholds were determined empirically by analyzing the dynamic range of motion magnitudes across the entire dataset.
\begin{itemize}
    \item \textbf{Strong Motion:} Average magnitude $>$ 15.0
    \item \textbf{Significant Motion:} Average magnitude $>$ 8.0
    \item \textbf{Subtle Motion:} Average magnitude $>$ 3.0
    \item \textbf{Micro-Motion:} Average magnitude $\leq$ 3.0
\end{itemize}
An ROI is considered to have "significant regional motion" if its average magnitude exceeds the 8.0 threshold.

\paragraph{Directional and Contraction Analysis.} For specific AUs, simple magnitude analysis is insufficient.
\begin{itemize}
    \item For upward movements like brow raisers (AU1, AU2), we employ a directional filter, only considering motion vectors within a $90^\circ$ arc centered on the vertical axis (i.e., between 45$^\circ$ and 135$^\circ$).
    \item For concentric actions like 'Lid Tightener' (AU7), we analyze the radial component of the flow vectors relative to the ROI's geometric center. An activation is confirmed if over 70\% of the vectors on the ROI boundary point inwards.
\end{itemize}

\subsubsection{Automated Rationale Generation}
The rationale ($\mathcal{R}$) for each sample is generated via a deterministic, rule-based system that translates the quantitative optical flow analysis into a structured, human-readable explanation. This system does not use a probabilistic language model, ensuring consistency and verifiability. The generation process follows a three-part structure:
\begin{enumerate}
    \item \textbf{Analysis Process:} This section provides a detailed, region-by-region summary of the observed facial motion. It describes the intensity and direction (represented by a color-and-text descriptor) of movement in regions corresponding to activated AUs. It also reports on any significant motion detected in non-activated regions, labeling it as potential interference, as identified by the dual-verification mechanism.
    \item \textbf{Expression Reasoning:} This part synthesizes the findings from the analysis into a coherent interpretation. It leverages a predefined knowledge base that maps combinations of AUs to prototypical emotional expressions (e.g., the co-occurrence of AU6 and AU12 is linked to happiness). It also comments on the symmetry or asymmetry of the expression based on a comparison of left- and right-side ROI activations.
    \item \textbf{Final Conclusion:} This section provides a definitive summary, listing the activated AUs and stating the final micro-expression category. Crucially, this emotion label is taken directly from the ground-truth annotation of the source dataset, serving as a verified conclusion rather than a prediction from our analysis system.
\end{enumerate}

\subsection{Data Statistics and Analysis}
The aggregation of multiple datasets introduces inherent imbalances in label distributions, which are important to characterize.

\subsubsection{Emotion Label Distribution}
The distribution of the eight emotion categories is presented in Table~\ref{tab:emotion_distribution}. While the six primary emotions (excluding 'Contempt' and 'Other') are relatively balanced (12-21\% each), the dataset is skewed towards negative emotions (Anger, Disgust, Fear, Sadness), which collectively account for 65.4\% of the samples. The 'Contempt' category is significantly underrepresented, comprising only 2.23\% of the data.

\begin{table}[ht!]
\centering
\caption{\textbf{Emotion Category Distribution in Uni-MER.} The dataset exhibits a relative balance among the six primary emotions, with a notable scarcity of the 'Contempt' class.}
\label{tab:emotion_distribution}
\resizebox{\columnwidth}{!}{%
\begin{tabular}{l|r|r}
\toprule
\textbf{Emotion Category} & \textbf{Sample Count} & \textbf{Percentage (\%)} \\
\midrule
Disgust        & 1,718 & 21.37\% \\
Fear           & 1,285 & 15.98\% \\
Anger          & 1,098 & 13.66\% \\
Happiness      & 1,058 & 13.16\% \\
Sadness        & 1,094 & 13.61\% \\
Surprise       & 957 & 11.91\% \\
Other          & 648   & 8.07\%  \\
Contempt       & 183   & 2.24\%  \\
\midrule
\textbf{Total} & \textbf{8,041} & \textbf{100.00\%} \\
\bottomrule
\end{tabular}%
}
\end{table}

\subsubsection{Action Unit (AU) Distribution}
The dataset exhibits a severe imbalance in AU labels (Table~\ref{tab:au_distribution}). The most frequent AUs, AU4 ('Brow Lowerer') and AU14 ('Dimpler'), appear in 33.3\% and 22.9\% of samples, respectively. In contrast, the least frequent AUs, AU9 ('Nose Wrinkler') and AU15 ('Lip Corner Depressor'), are present in fewer than 2.5\% of samples. This long-tail distribution poses a significant challenge for models intended to recognize the full spectrum of AUs.

\begin{table}[ht!]
\centering
\caption{\textbf{Action Unit (AU) Label Distribution in Uni-MER.} The occurrence of AUs is highly imbalanced, with AU4 and AU14 being far more prevalent than others.}
\label{tab:au_distribution}
\resizebox{\columnwidth}{!}{%
\begin{tabular}{l|l|r|r}
\toprule
\textbf{AU} & \textbf{Description} & \textbf{Count} & \textbf{Percentage (\%)} \\
\midrule
AU4         & Brow Lowerer         & 2,678 & 33.30\% \\
AU14        & Dimpler              & 1,845 & 22.94\% \\
AU1         & Inner Brow Raiser    & 1,165 & 14.49\% \\
AU12        & Lip Corner Puller    & 1,074 & 13.36\% \\
AU2         & Outer Brow Raiser    & 1,068 & 13.28\% \\
AU7         & Lid Tightener        & 1,008 & 12.54\% \\
AU5         & Upper Lid Raiser     & 668   & 8.31\%  \\
AU6         & Cheek Raiser         & 431   & 5.36\%  \\
AU10        & Upper Lip Raiser     & 348   & 4.33\%  \\
AU17        & Chin Raiser          & 333   & 4.14\%  \\
AU9         & Nose Wrinkler        & 191   & 2.37\%  \\
AU15        & Lip Corner Depressor & 183   & 2.28\%  \\
\bottomrule
\end{tabular}%
}
\end{table}
\begin{table}[t!]
  \centering
  \caption{The following is a sample prompt from our \texttt{[emotion]} instruction pool, guiding the model to perform process, reasoning, and conclusion steps, and return results in structured JSON.}
  \label{tab:emotion_instructions}
  \vspace{3mm}
  \fcolorbox{black}{gray!15}{%
    \begin{minipage}{0.9\linewidth}
      \small
      \begin{itemize}
          \setlength\itemsep{0.5em}
          \item Return JSON only with reasoning: 
          
          \item "emotion": one of [anger, sadness, happiness, fear, disgust, surprise, other, contempt], 
          \item "rationale": Analyze the optical flow visualization where hue indicates motion direction and brightness indicates motion intensity. The color-to-direction map is: Up=purple, Upper-Right=pink, Right=red, Lower-Right=orange, Down=yellow-green, Lower-Left=green, Left=cyan, Upper-Left=blue. Structure your response with these headings:

          \begin{enumerate}
            \item Analysis process: Based on the color blocks, identify the active facial regions and their motion patterns. Focus on eyebrow movements (raising/lowering), eye movements (lid changes), nose/upper lip changes, and mouth/chin movements.
            \item  Expression reasoning: Explain how the combination of these facial movements suggests a specific emotion.
            \item Conclusion: State the final emotion, choosing from: anger, sadness, happiness, fear, disgust, surprise, other, contempt.
          \end{enumerate}
         
      \end{itemize}
    \end{minipage}%
  }
\end{table}

\subsubsection{Source Dataset vs. Emotion Distribution}
As shown in Table~\ref{tab:crosstab_dataset_emotion}, the emotional content varies significantly across the source datasets. For instance, the 'Contempt' emotion is almost exclusively present in the DFME series and SAMM datasets. Similarly, CASME3\_MaE is particularly rich in 'Fear' and 'Sadness' samples, while the DFME datasets contribute a large number of 'Disgust' samples. This cross-source diversity is a key strength of Uni-MER, as it ensures that models trained on it are exposed to varied emotional expressions from different populations and elicitation contexts.

\begin{table*}[ht!]
\centering
\caption{\textbf{Cross-Tabulation of Emotion Categories by Source Dataset.} This table highlights the complementary nature of the source datasets, with different datasets contributing varied distributions of emotional expressions.}
\label{tab:crosstab_dataset_emotion}
\resizebox{\textwidth}{!}{%
\begin{tabular}{l|rrrrrrrr|r}
\toprule
\textbf{Dataset} & \textbf{Happiness} & \textbf{Sadness} & \textbf{Surprise} & \textbf{Fear} & \textbf{Anger} & \textbf{Disgust} & \textbf{Contempt} & \textbf{Other} & \textbf{Total} \\
\midrule
CASME           & 7                  & 6                & 21                & 2             & 0              & 44               & 0                 & 108            & 188            \\
$\mathrm{CAS(ME)^2}$           & 147                & 8                & 36                & 21            & 55             & 74               & 0                 & 13             & 354            \\
$\mathrm{CAS(ME)^3}$         & 54                 & 57               & 185               & 85            & 61             & 248              & 0                 & 150            & 840            \\
$\mathrm{CAS(ME)^3\_MaE}$    & 367                & 642              & 80                & 787           & 487            & 475              & 0                 & 107            & 2,945          \\
CASME II         & 32                 & 4                & 28                & 2             & 0              & 63               & 0                 & 126            & 255            \\
D4ME            & 56                 & 0                & 32                & 0             & 140            & 0                & 0                 & 37             & 265            \\
D4ME\_MAE       & 23                 & 0                & 25                & 0             & 50             & 0                & 0                 & 17             & 115            \\
DFME            & 206                & 278              & 298               & 265           & 161            & 548              & 100               & 0              & 1,856          \\
DFME\_TEST\_A     & 63                 & 46               & 101               & 62            & 39             & 129              & 34                & 0              & 474            \\
DFME\_TEST\_B     & 42                 & 35               & 48                & 38            & 41             & 58               & 37                & 0              & 299            \\
MMEW            & 35                 & 12               & 88                & 15            & 8              & 70               & 0                 & 64             & 292            \\
SAMM            & 26                 & 6                & 15                & 8             & 56             & 9                & 12                & 26             & 158            \\
\midrule
\textbf{Total}  & \textbf{1,058}     & \textbf{1,094}   & \textbf{957}    & \textbf{1,285} & \textbf{1,098} & \textbf{1,718}   & \textbf{183}      & \textbf{648}   & \textbf{8,041} \\
\bottomrule
\end{tabular}%
}
\end{table*}
\begin{table}[ht!]
  \centering
  \caption{The universal system prompt used for all experiments. It defines the model's expert persona, its operational rules, and its constraints regarding multimodal evidence and output generation.}
  \label{tab:system_prompt}
  \vspace{3mm}
  \fcolorbox{black}{gray!15}{%
    \begin{minipage}{0.9\linewidth}
      \small
      You are a micro-expression analysis assistant specialized in linking Facial Action Units (AUs) to emotions and inferring AUs from motion cues in optical flow under user-provided conventions.
      
      \textbf{General rules:}
      \begin{itemize}
          \setlength\itemsep{0.2em}
          \item Follow the dialogue instructions strictly; reason stepwise but only output final structured findings.
          \item Treat \texttt{<feature>} contents as evidence containers provided by the user; do not hallucinate missing signals.
      \end{itemize}
      
      \textbf{Multimodal feature policy (text-aligned features):}
      \begin{itemize}
          \setlength\itemsep{0.2em}
          \item Inputs may include features produced by an external image/video encoder and projected into the language space; treat tokens in \texttt{<feature>} as aligned, high-priority cues but still cross-check against context.
          \item Use internal knowledge of AU–emotion relationships, but defer to explicit evidence when conflicts occur.
          \item Assume the optical-flow hue/direction convention and any AU region-of-interest definition are provided in the user message; do not invent unmapped conventions.
      \end{itemize}
      
      \textbf{Constraints:}
      \begin{itemize}
          \setlength\itemsep{0.2em}
          \item No medical or legal claims; calibrate uncertainty; remain concise and factual.
          \item If features are missing, say so and proceed with what is available.
      \end{itemize}
    \end{minipage}%
  }
\end{table}
\section{Multimodal Prompt Templates}
\label{sec:appendix_d}

This appendix provides the complete and unabridged prompt templates used for our multimodal micro-expression analysis, as referenced in Sec. \ref{sec:3.3}. The detailed specification of these templates is provided to ensure full reproducibility of our experiments. Our prompting strategy is designed to be robust, task-specific, and capable of eliciting structured, reasoned outputs from the large language model.

\subsection{Prompt Architecture}
\label{sec:prompt_architecture}

All interactions with the model follow a consistent high-level architecture. The final input string is a concatenation of a global system prompt, multimodal placeholders for visual data, and a task-specific user prompt. The structure is as follows:
\begin{center}
[\textit{System Prompt}] \\ \textit{\textless feature\textgreater\textless ExpertFeatureHere\textgreater\textless /feature\textgreater} \\
\textit{[Task Identifier] Prompt}
\end{center}
\begin{itemize}
    \item \textbf{System Prompt:} A global set of instructions defining the model's persona, capabilities, and constraints. This prompt is constant across all tasks.
    \textbf{\texttt{<ExpertFeatureHere>}} placeholders: These are reserved tokens where the  extracted multimodal features are injected.
    \item \textbf{Task Identifier:} A special token (\texttt{[emotion]} or \texttt{[flow]}) that precedes the user prompt to prime the model for a specific task.
    \item \textbf{User Prompt:} A detailed instruction selected from a task-specific pool, which guides the model's analysis and specifies the desired output format.
\end{itemize}

\subsection{System Prompt}
The universal system prompt, detailed in Table~\ref{tab:system_prompt}, establishes the model's role as a specialized micro-expression analysis assistant. It outlines general rules, policies for handling multimodal features, and strict operational constraints.

\subsection{Task-Specific Instruction Pools}
To enhance model robustness and prevent overfitting to a single phrasal structure, we employ instruction pools for each task. During training and inference, a prompt is randomly selected from the corresponding pool.

\subsubsection{Emotion Recognition Task ([emotion])}
The instructions in this pool (Table~\ref{tab:emotion_instructions}) direct the model to analyze the provided optical flow visualization and infer one of eight emotion categories. Each prompt variant requests the same core analysis but with slightly different wording and structure.

\subsubsection{Action Unit (AU) Recognition Task ([flow])}
This instruction pool (Table~\ref{tab:flow_instructions}) guides the model to identify active AUs from the optical flow visualization. The prompts vary in their level of guidance, from providing explicit AU knowledge to requesting a general analysis based on FACS principles.

\begin{table}[t!]
  \centering
  \caption{The representative example from the instruction pool for the Action Unit (AU) recognition task (\texttt{[flow]}). Each prompt instructs the model to return a JSON object specifying the detected AUs, corresponding evidence, and a rationale for its predictions.}
  \label{tab:flow_instructions}
  \vspace{3mm}
  \fcolorbox{black}{gray!15}{%
    \begin{minipage}{0.9\linewidth}
      \small
      \begin{itemize}
          \setlength\itemsep{0.5em}
          \item Using FACS principles, analyze facial motion patterns in the optical flow map. Key regions: eyebrows (brow raising/lowering), eyes (lid movements), nose (nasal changes), mouth (lip movements). Focus on bilateral symmetry and motion intensity. Ignore weak signals and lighting artifacts. Return JSON:
          \item  "aus":  \texttt{["AUXX","AUYY",...]}, 
          \item "evidence": \texttt{"region name": \{"direction": "Up|UR|R|LR|D|LL|L|UL", "intensity": "low|medium|high"\}}, 
          \item "rationale":\texttt{\{ "FACS-based systematic analysis"\}.'}
      \end{itemize}
    \end{minipage}%
  }
\end{table}

\subsection{Optical Flow Visualization Convention}
All user prompts for both tasks reference a consistent convention for interpreting the optical flow visualizations, as summarized in Table~\ref{tab:flow_convention}. In this convention, the hue of a pixel encodes the direction of motion, while its brightness (value) encodes the intensity (magnitude) of the motion.

\begin{table}[ht!]
\centering
\caption{Color-to-direction mapping convention for optical flow visualizations.}
\label{tab:flow_convention}
\begin{tabular}{l|l|l}
\toprule
\textbf{Color} & \textbf{Direction Code} & \textbf{Description} \\
\midrule
Purple & Up / U & Upward motion \\
Pink / Magenta & Upper-Right / UR & Upper-right motion \\
Red & Right / R & Rightward motion \\
Orange & Lower-Right / LR & Lower-right motion \\
Yellow-Green & Down / D & Downward motion \\
Green & Lower-Left / LL & Lower-left motion \\
Cyan & Left / L & Leftward motion \\
Blue & Upper-Left / UL & Upper-left motion \\
\bottomrule
\end{tabular}
\end{table}

\section{Hyperparameter Details}
\label{sec:appendix_e}

This appendix provides a comprehensive specification of the hyperparameters used for training our model, as referenced in Sec. \ref{sec:4.1}. All parameters are detailed here to ensure the full reproducibility of our experimental results.

\subsection{Main Model Optimization}
\paragraph{Optimizer.} We employ the AdamW optimizer~\cite{loshchilov2019decoupled} for training the primary model. It is configured with a $\beta_1$ of 0.9, a $\beta_2$ of 0.999, and a weight decay of 0.05.

\paragraph{Learning Rate Schedule.} A linear warm-up followed by a cosine annealing schedule is used. The learning rate begins at $1 \times 10^{-5}$, warms up to a peak of $1 \times 10^{-4}$ over 500 steps, and then decays to a minimum of $1 \times 10^{-5}$ over 40 epochs (1,000 iterations per epoch).

\subsection{Expert Encoder Fine-Tuning}
The Motion-Semantics Expert ($\mathcal{E}_{sem}$) undergoes a separate fine-tuning stage prior to the main model training. The hyperparameters for this stage are:
\begin{itemize}
    \item \textbf{Optimizer:} AdamW
    \item \textbf{Learning Rate:} $1 \times 10^{-4}$ (constant)
    \item \textbf{Batch Size:} 16
    \item \textbf{Epochs:} 100
\end{itemize}

\subsection{Loss Formulation and Weighting}
The total loss is a weighted sum of multiple components. The primary generative text loss is weighted by $w_{evi} = 0.6$. The auxiliary AU classification loss is weighted by $w_{au} = 1.2$, and the emotion classification loss is weighted by $w_{emo} = 1.2$. Rationale-specific samples receive an additional weight of $w_{rat} = 0.3$.

\pagebreak[4]
\paragraph{AU Classification Loss.} We use a Focal Loss with a focusing parameter $\gamma=2.0$ for the multi-label AU classification task.
\paragraph{Token-Level Weighting.} For the emotion recognition task, the tokens corresponding to the final emotion label in the generative output are assigned a weight of 2.0, while all other rationale tokens have a weight of 0.3.

\subsection{Dataset Splits for Experiments}
The training and validation splits were configured based on the experimental setup in Sec. \ref{sec:4.2}.
\begin{enumerate}
    \item \textbf{Unseen AU Combination Setting:} Four source datasets (CAS(ME)$^3$\_MaE, DFME, CAS(ME)$^2$, D4ME MAE) were used for training, with six held out for validation.
    \item \textbf{Leave-One-Dataset-Out Setting:} 10 of the 12 source datasets were used for training, holding out one for testing in a leave-one-out scheme.
\end{enumerate}
In all experiments, the DFME TEST\_A and DFME TEST\_B datasets were exclusively reserved for testing.

\subsection{Hyperparameter Summary Table}
A consolidated list of all key hyperparameters is provided in Table~\ref{tab:hyperparameters}.
\begin{table*}[t!]
\centering
\caption{Consolidated table of all major hyperparameters used for training and model configuration.}
\label{tab:hyperparameters}
\resizebox{\textwidth}{!}{%
\begin{tabular}{l|l|l|l}
\toprule
\textbf{Category} & \textbf{Parameter} & \textbf{Value} & \textbf{Description} \\
\midrule
\hline
\multicolumn{4}{c}{\textbf{Optimizer \& Learning Rate}} \\
\hline
Optimizer & Type & AdamW & The optimization algorithm used. \\
& Learning Rate (Initial) & $1 \times 10^{-4}$ & The peak learning rate after warm-up. \\
& Learning Rate (Minimum) & $1 \times 10^{-5}$ & The target learning rate after cosine decay. \\
& Learning Rate (Warm-up) & $1 \times 10^{-5}$ & The starting learning rate for the warm-up phase. \\
& Weight Decay & 0.05 & The coefficient for L2 regularization. \\
& Beta Coefficients & (0.9, 0.999) & AdamW $\beta_1$ and $\beta_2$ parameters. \\
& Epsilon ($\epsilon$) & $1 \times 10^{-8}$ & Term for numerical stability (PyTorch default). \\
LR Scheduler & Type & Linear Warm-up + Cosine Annealing & The learning rate scheduling strategy. \\
& Warm-up Steps & 500 & Number of initial steps for linear LR increase. \\
& Max Epochs & 40 & Total number of training epochs. \\
& Iterations per Epoch & 1,000 & Number of training steps within a single epoch. \\
\hline
\multicolumn{4}{c}{\textbf{Model Architecture \& LoRA}} \\
\hline
LLM & Base Model & Meta-Llama-3.1-8B-Instruct & The foundational large language model. \\
LoRA & Rank ($r$) & 64 & The rank of the decomposition matrices. \\
& Alpha ($\alpha$) & 16 & The scaling factor for LoRA. \\
& Dropout & 0.05 & Dropout probability applied to LoRA matrices. \\
& Target Modules & [q\_proj, k\_proj, v\_proj, o\_proj, gate\_proj, up\_proj, down\_proj] & LLM layers adapted with LoRA. \\
Projection Layers & Architecture & Single Linear Layer & All five modality-bridging projectors. \\
& $\sigma_{temp}, \sigma_{struc}, \sigma_{sem}$ & Linear(1024 $\to$ 4096) & Projects expert features to LLM space. \\
DCM Heads & Architecture & Single Linear Layer & Both auxiliary classification heads. \\
& AU Head Output & 12 & Number of Action Unit classes. \\
& Emotion Head Output & 8 & Number of emotion classes. \\
\hline
\multicolumn{4}{c}{\textbf{Loss Function}} \\
\hline
Loss Weights & Generative Evidence ($w_{evi}$) & 0.6 & Weight for the main text generation loss. \\
& AU Classification ($w_{au}$) & 1.2 & Weight for the auxiliary AU classification loss. \\
& Emotion Classification ($w_{emo}$) & 1.2 & Weight for the auxiliary emotion classification loss. \\
& Rationale Samples ($w_{rat}$) & 0.3 & Additional weight for rationale-annotated samples. \\
AU Loss & Type & Focal Loss & Loss function for the AU head. \\
& Gamma ($\gamma$) & 2.0 & The focusing parameter for the Focal Loss. \\
Emotion Loss & Type & Cross-Entropy & Loss function for the emotion head. \\
Token Weighting & Emotion Label Token & 2.0 & Weight for the final emotion token in the generative loss. \\
& Default Token & 0.3 & Weight for all other non-label tokens. \\
\hline
\multicolumn{4}{c}{\textbf{Training Strategy \& Data}} \\
\hline
Training & Batch Size & 1 & Number of samples per forward/backward pass. \\
& Sample Ratio & 30 & Controls data sampling frequency per epoch. \\
& Data Workers & 6 & Number of parallel processes for data loading. \\
& Mixed Precision (AMP) & True (FP16) & Enables half-precision training. \\
& Random Seed & 42 & Seed for all random number generators. \\
& Gradient Checkpointing & True & Reduces memory usage at the cost of computation. \\
Parallelism & Strategy & Pipeline Parallelism & Method for distributing model across multiple GPUs. \\
& Devices & 4 GPUs & Total number of GPUs used for a single model. \\
& Layer Distribution & [2, 10, 10, 10] & Distribution of LLM layers across the 4 GPUs. \\
Data Specs & Image Size & 224 $\times$ 224 & The spatial resolution for all visual inputs. \\
& Max Text Length & 1024 & Maximum number of tokens in any text sequence. \\
\bottomrule
\end{tabular}%
}
\end{table*}

\section{Architectural Details}
\label{sec:appendix_f}

This appendix details the architectural specifications of our model's core components, including the expert encoders, feature projectors, and the discriminative calibration module.

\subsection{Expert Encoders}
Our model leverages three specialized, pre-trained expert encoders to extract rich multimodal features.

\paragraph{Temporal Expert ($\mathcal{E}_{temp}$).} This expert utilizes a VideoMAE \cite{tong2022videomae} encoder to model temporal dynamics. To capture the full evolution of a micro-expression, which is inherently brief, we do not subsample frames. Instead, the entire sequence from the onset to the offset frame is fed to the encoder. This ensures that subtle, transient motion patterns are not lost. The encoder processes these frames to produce a 1024-dimensional feature vector representing the video's temporal signature.

\paragraph{Structural Expert ($\mathcal{E}_{struc}$).} To capture facial structure and its relation to expressive regions, we adapt the methodology from~\cite{zheng2023poster}. This expert employs a cross-attention mechanism to fuse landmark and image information. The query ($Q$) is derived from the final feature layer of a MobileFaceNet model responsible for landmark detection. The key ($K$) and value ($V$) are derived from the final layer of an IR-50 backbone, which extracts deep facial features. The module is pre-trained using both AU and emotion labels, following the same data partitioning strategy as the main LLM, to learn a structural representation of the face that is highly correlated with expressions. The output is a 1024-dimensional feature vector.

\paragraph{Motion-Semantics Expert ($\mathcal{E}_{sem}$).} This expert is designed to bridge the gap between low-level motion (optical flow) and high-level semantics (AUs and emotions). We start with a pre-trained SigLIP model (\texttt{siglip-so400m-patch14-224}). While SigLIP's contrastive pre-training provides a strong baseline for vision-language alignment, we specialize it for our domain. We fine-tune the upper 50\% of the SigLIP vision encoder's layers using LoRA. During this fine-tuning stage, two linear classification heads are attached to the encoder's output: one for AU classification and one for emotion classification. By training on AU and emotion labels, the encoder learns to produce a 1024-dimensional feature vector that is not just visually descriptive but also semantically aligned with the specific categories of our micro-expression tasks.

\subsection{Feature Projection and Adaptation}
\paragraph{Projection Layers ($\sigma$).} The features from all three expert encoders are projected into the LLM's latent space using lightweight, single-layer linear projectors. Each projection layer ($\sigma_{temp}$, $\sigma_{struc}$, and $\sigma_{sem}$) maps its corresponding 1024-dimensional expert feature vector to a 4096-dimensional vector, matching the hidden dimension of the LLaMA-3.1-8B model.

\paragraph{LoRA Adaptation.} The LLaMA-3.1-8B model is adapted using LoRA. The target modules for this adaptation are the linear layers within the self-attention and feed-forward blocks: \texttt{q\_proj, k\_proj, v\_proj, o\_proj, gate\_proj, up\_proj,} and \texttt{down\_proj}.

\subsection{Discriminative Calibration Module (DCM)}
\paragraph{Architecture.} The DCM is composed of two parallel, single-layer linear classifiers that function as auxiliary heads for AU and emotion prediction, respectively. These heads do not contain any intermediate non-linearities or hidden layers.

\paragraph{Input Feature Derivation.} The input to the DCM is a single 4096-dimensional vector derived via a semantic pooling mechanism. This mechanism operates on the final hidden states of the LLaMA model. Specifically, it exclusively considers the hidden state representations corresponding to the locations of the injected expert prefix tokens ($\mathcal{T}_{deft}$). The hidden states at all other token positions are masked out. The selected hidden states are then average-pooled to produce a single feature vector, which is then fed concurrently to both the AU and emotion classification heads of the DCM. This ensures that the discriminative task is performed only on the basis of the multimodal evidence provided by the expert encoders.

\section{Additional Ablation Studies}
\label{sec:appendix_g}

\subsection{Contribution of Expert Encoders}
To investigate the individual and synergistic contributions of our three expert encoders, we conducted an incremental ablation study. The Motion-Semantics Expert ($\mathcal{E}_{sem}$) serves as the foundational component, as its role in bridging optical flow with textual semantics is critical for the model's basic convergence. We then progressively add the Temporal Expert ($\mathcal{E}_{temp}$) and the Structural Expert ($\mathcal{E}_{struc}$). The results are presented in Table~\ref{tab:ablation_experts}.

\begin{table}[ht!]
\centering
\caption{Ablation study on the contribution of expert encoders. The Motion-Semantics Expert ($\mathcal{E}_{sem}$) is maintained as the essential baseline. All metrics are reported in percent (\%).}
\label{tab:ablation_experts}
\resizebox{\columnwidth}{!}{%
\begin{tabular}{ccc|ccc|ccc}
\toprule
\multicolumn{3}{c|}{\textbf{Expert Modules}} & \multicolumn{3}{c|}{\textbf{DFME\_TEST\_A}} & \multicolumn{3}{c}{\textbf{DFME\_TEST\_B}} \\
$\mathcal{E}_{sem}$ & $\mathcal{E}_{temp}$ & $\mathcal{E}_{struc}$ & UF1 & UAR & ACC & UF1 & UAR & ACC \\
\midrule
\checkmark & & & 37.07 & 39.01 & 47.05 & 36.44 & 36.11 & 37.79 \\
\checkmark & \checkmark & & 39.45 & 41.15 & 49.58 & 39.61 & 40.00 & 42.47 \\
\checkmark & \checkmark & \checkmark & \textbf{43.72} & \textbf{42.13} & \textbf{51.26} & \textbf{42.81} & \textbf{42.24} & \textbf{43.47} \\
\bottomrule
\end{tabular}%
}
\end{table}

\paragraph{Analysis.}
The results clearly demonstrate the complementary nature of the expert encoders.

\textbf{Baseline ($\mathcal{E}_{sem}$ only):} The model equipped solely with the Motion-Semantics Expert establishes a reasonable performance baseline (e.g., 37.07\% UF1 on Test A). This confirms that grounding the model in the semantic meaning of motion is fundamental. Without this expert, the model fails to map the raw pixel displacements from optical flow to a meaningful conceptual space, leading to a failure to converge. $\mathcal{E}_{sem}$ provides the essential "what"---translating motion patterns into semantic concepts like "brow lowering."

\textbf{Adding Temporal Context ($\mathcal{E}_{sem} + \mathcal{E}_{temp}$):} The introduction of the Temporal Expert yields a significant performance improvement across all metrics (e.g., UF1 on Test A increases by 2.38\% to 39.45\%). This highlights the importance of understanding the evolution of a micro-expression over time. While $\mathcal{E}_{sem}$ identifies the motion, $\mathcal{E}_{temp}$ provides the temporal context---the "how"---allowing the model to distinguish between subtle expressions based on their unique dynamic signatures, such as the speed and progression of muscle activation.

\textbf{Adding Structural Grounding ($\mathcal{E}_{sem} + \mathcal{E}_{temp} + \mathcal{E}_{struc}$):} The inclusion of the Structural Expert brings a further substantial boost in performance, leading to our final model's results (e.g., UF1 on Test A increases by another 4.27\% to 43.72\%). This demonstrates the value of anatomically grounding the motion analysis. $\mathcal{E}_{struc}$ provides a strong spatial prior by fusing landmark information with facial appearance, answering "where" the motion is occurring. This helps the model to disambiguate motion patterns (e.g., differentiating a cheek raise from a lip corner pull) and focus its analysis on facially relevant regions, making the final inference more robust and accurate.

Collectively, these results validate our three-expert design, showing that semantic, temporal, and structural information are all critical and non-redundant for high-performance micro-expression analysis.

\begin{figure*}[ht!]
    \centering
    \includegraphics[width=\textwidth]{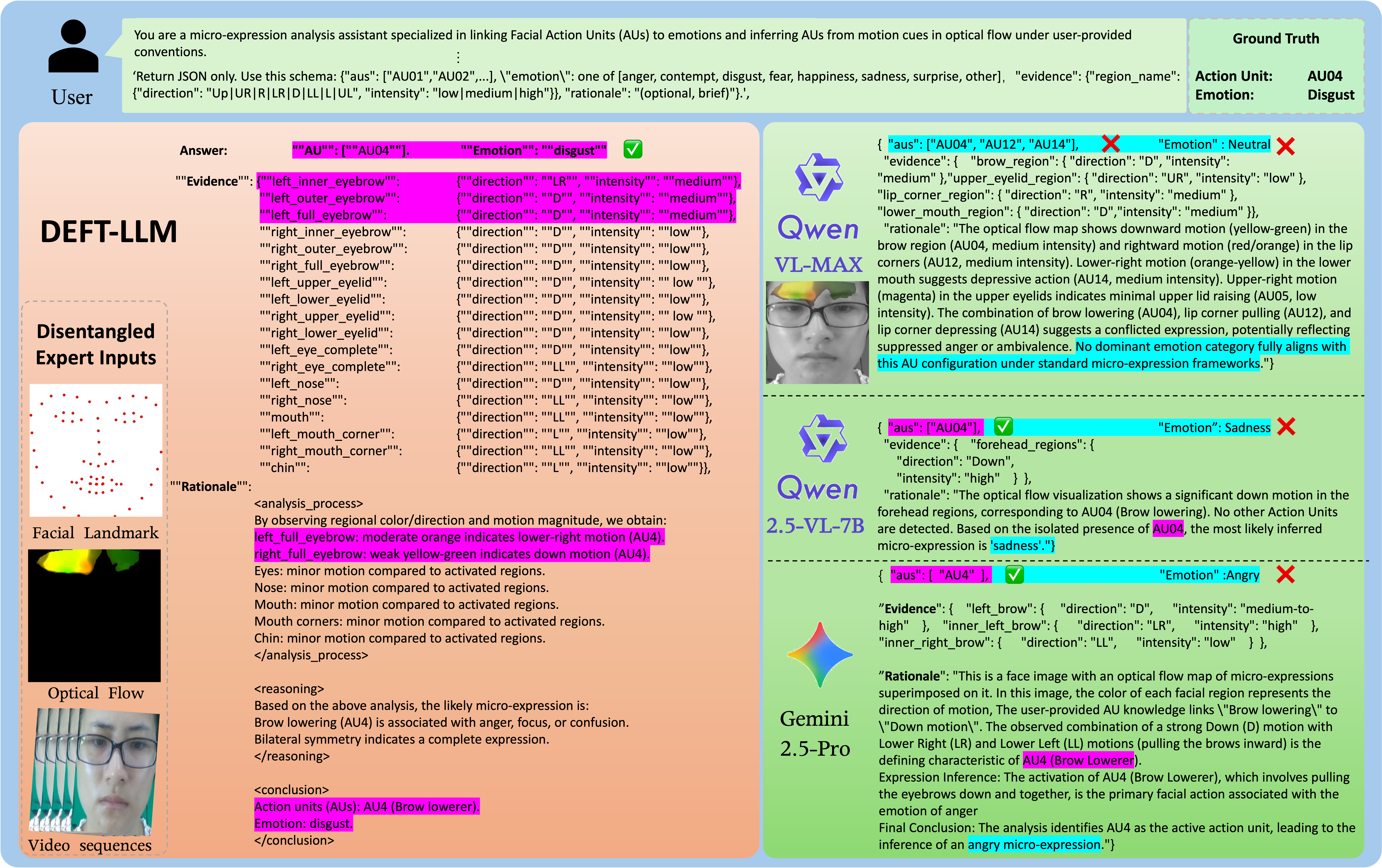} 
    \caption{\textbf{Qualitative comparison on a sample with Ground Truth AU4 and Disgust.} Our model, DEFT-LLM, accurately identifies both the Action Unit and the subtle emotion, providing a detailed and correct rationale. In contrast, generalist MLLMs like Qwen-VL-MAX, Qwen-2.5-VL-7B, and Gemini 2.5-Pro exhibit various failure modes. This highlights the necessity of feature disentanglement for the nuanced task of micro-expression recognition.Detailed analysis can be found in Appendix \ref{sec:appendix_h}}.
    \label{fig:qualitative_comparison}
\end{figure*}

\section{Qualitative Results}
\label{sec:appendix_h}

To complement the quantitative metrics presented in the main paper, this section provides a qualitative case study. This analysis serves to visually and textually demonstrate our model's advanced capabilities in interpretability and accuracy compared to leading generalist Multimodal Large Language Models (MLLMs). We present a challenging case where the ground truth is Action Unit AU4 ('Brow Lowerer') and the emotion is 'Disgust'. The comparative results are shown in Figure~\ref{fig:qualitative_comparison}.

\subsection{DEFT-LLM: Accurate and Interpretable Analysis}
As shown in Figure~\ref{fig:qualitative_comparison}, our DEFT-LLM correctly identifies both the active Action Unit as AU4 and the emotion as 'Disgust'. The model's generated 'Evidence' meticulously details the motion observed across all facial regions, correctly pinpointing downward motion ("D") of medium intensity in the 'left full eyebrow' and 'right full eyebrow' regions.

The 'Rationale' demonstrates a sound reasoning process. It first analyzes the visual evidence, then correctly associates the bilateral brow lowering with AU4, and finally concludes that this specific activation pattern corresponds to 'Disgust'. This success is not merely a correct prediction but a demonstration of correct reasoning, which stems directly from our model's architecture:
\begin{enumerate}
    \item \textbf{Decoupled Expertise:} The three expert encoders provide distinct, uncorrupted signals representing motion semantics, temporal dynamics, and facial structure. The model is not forced to interpret a convoluted, entangled representation of RGB and flow.
    \item \textbf{Informed Synthesis:} The LLM leverages its vast prior knowledge to synthesize these discrete clues. It understands that while AU4 can be associated with several negative emotions, the specific context provided by the other experts helps to disambiguate it, correctly inferring 'Disgust' over the more generic 'Anger' or 'Sadness'. The model learns the implicit connections between the multi-expert clues and the ground truth.
\end{enumerate}
\subsection{Comparative Analysis with Generalist MLLMs}
The limitations of general-purpose MLLMs become apparent when faced with the subtlety of micro-expressions.

\paragraph{Qwen-VL-MAX: Evidential Hallucination and Misinterpretation.}
This model exhibits a critical failure mode: hallucination. It incorrectly identifies AU12 ('Lip Corner Puller') and AU14 ('Dimpler'), motions associated with smiling. This is the diametrical opposite of the subtle downward lip movement observable in the video. This error likely arises from the model's attempt to interpret an entangled representation of RGB and flow. Lacking a specialized understanding of micro-expressions, it may default to recognizing more common, high-energy macro-expressions, leading it to misinterpret subtle negative cues as components of a smile.

\paragraph{Qwen-2.5-VL-7B and Gemini 2.5-Pro: Oversimplification and Lack of Specificity.}
These models successfully identify the most salient visual cue: the downward brow motion corresponding to AU4. This demonstrates a basic capability in motion perception. However, their failure lies in the subsequent reasoning step. They map AU4 to 'Sadness' (Qwen) and 'Anger' (Gemini), respectively. While these are plausible associations for AU4 in a general context, they are incorrect for this specific instance.

This highlights a core challenge in MER that these models fail to address: the one-to-many mapping of AUs to emotions. A single AU rarely defines an emotion uniquely. The models' failures suggest an over-reliance on a simplistic, direct mapping from a single, dominant visual feature (optical flow) to an emotion category. They lack the multi-faceted context provided by our decoupled experts, which is necessary to discern the subtle differences between disgust, anger, and sadness that all might share a common AU4 component.

\subsection{Conclusion of Qualitative Analysis}
This case study underscores that effective micro-expression recognition requires more than just accurate motion detection. The entanglement of RGB and flow features in generalist models can corrupt the subtle information present in the original image, leading to misinterpretation and hallucination. In contrast, our decoupled, multi-expert approach provides the LLM with clean, specialized clues. This allows the model to leverage its powerful reasoning capabilities and vast prior knowledge to accurately interpret the complex, often ambiguous, nature of micro-expressions, validating the design philosophy behind DEFT-LLM.

\end{document}